\documentclass[acmtog, nonacm]{acmart}

\setcopyright{none}

\makeatletter
\let\@authorsaddresses\@empty       
\def\@mkauthorsaddresses{}          
\let\footnotetextauthorsaddresses\@gobble 
                                          
\makeatother

\makeatletter
\let\@mktitle\@mktitle@iii
\renewcommand{\@mkauthors}{\begingroup
  \hsize=\textwidth
  \global\setbox\mktitle@bx=\vbox{%
    \noindent\unvbox\mktitle@bx
    \begin{center}%
      {\Large\lineskip 0.5em%
        Sai Kumar Dwivedi\textsuperscript{1,2,\,$\dagger$}\quad
        Federica Bogo\textsuperscript{1}\quad
        Bu\u{g}ra Tekin\textsuperscript{1}\quad
        Chenhongyi Yang\textsuperscript{1}\\[2pt]
        Nadine Bertsch\textsuperscript{1}\quad
        Tomas Hodan\textsuperscript{1}\quad
        Michael J.\ Black\textsuperscript{2}\quad
        Dimitrios Tzionas\textsuperscript{3,4}\quad
        Shreyas Hampali\textsuperscript{1}\par}%
      \vspace{12pt}%
      {\normalsize
        \textsuperscript{1}Meta, Switzerland\quad
        \textsuperscript{2}Max Planck Institute for Intelligent Systems, T\"ubingen, Germany\\
        \textsuperscript{3}University of Amsterdam, Netherlands\quad
        \textsuperscript{4}Aristotle University of Thessaloniki, Greece\par}%
    \end{center}%
  }%
  \endgroup
}
\makeatother

\newcommand{\method}{\textsc{IMAGIN-4D}\xspace}

\usepackage{graphicx}
\usepackage{amsmath}
\usepackage{booktabs}
\usepackage{blindtext}

\definecolor{citecolor}{HTML}{0071bc}
\definecolor{frontcolor}{HTML}{325ea5}
\definecolor{sidecolor}{HTML}{a58b77}
\definecolor{DeltaColor}{rgb}{0.039,0.73,0.71}
\definecolor{SigmaColor}{rgb}{0.98,0.45,0.0}
\definecolor{AlphaColor}{rgb}{0,0,0.8}
\definecolor{BetaColor}{rgb}{0.8,0,0.8}
\definecolor{GammaColor}{rgb}{0.514,0.34,0.224}
\definecolor{EpsilonColor}{rgb}{0.353,0.725,0.906}
\definecolor{PurpleColor}{HTML}{bca5ea}
\definecolor{OrangeColor}{rgb}{0.914,0.541,0.0.141}
\definecolor{GreenColor}{rgb}{0.137,0.573,0.565}
\definecolor{RedColor}{rgb}{0.949,0.275, 0.224}
\definecolor{LightCyan}{rgb}{0.88,1,1}
\definecolor{Gray}{gray}{0.85}
\definecolor{LilacColor}{HTML}{8D538D} %

\newcommand{\TODO}[1]{}

\newcommand{\zheading}[1]{\hspace*{\parindent}\textbf{#1.}\ }
\newcommand{\qheading}[1]{\noindent\textbf{#1.}\ }

\newcommand{\supref}[1]{supmat\xspace}
\newcommand{\supmat}[1]{Sup.~Mat.}%

\usepackage{amsmath}
\usepackage{booktabs}
\usepackage{enumitem}
\usepackage{graphicx}
\usepackage{multirow}
\usepackage{tabularx}
\usepackage{xspace}
\usepackage{makecell}
\usepackage{pifont}

\usepackage{tabularx}
\usepackage{makecell}
\usepackage{booktabs}

\usepackage[normalem]{ulem} %

\usepackage{placeins}
\usepackage{lipsum}

\makeatletter
\@ifpackageloaded{hyperref}{}{%
  \definecolor{cvprblue}{rgb}{0.21,0.49,0.74}%
  \usepackage[pagebackref,breaklinks,colorlinks,citecolor=cvprblue]{hyperref}%
}
\makeatother

\begin{document}

\title{\textsc{IMAGIN-4D}: Image-Guided Controllable Interaction Generation}

\renewcommand{\shortauthors}{Dwivedi et al.}
\renewcommand{\shorttitle}{\textsc{IMAGIN-4D}: Image-Guided Controllable Interaction Generation}

\begin{teaserfigure}
\centering
\includegraphics[width=1.0\textwidth]{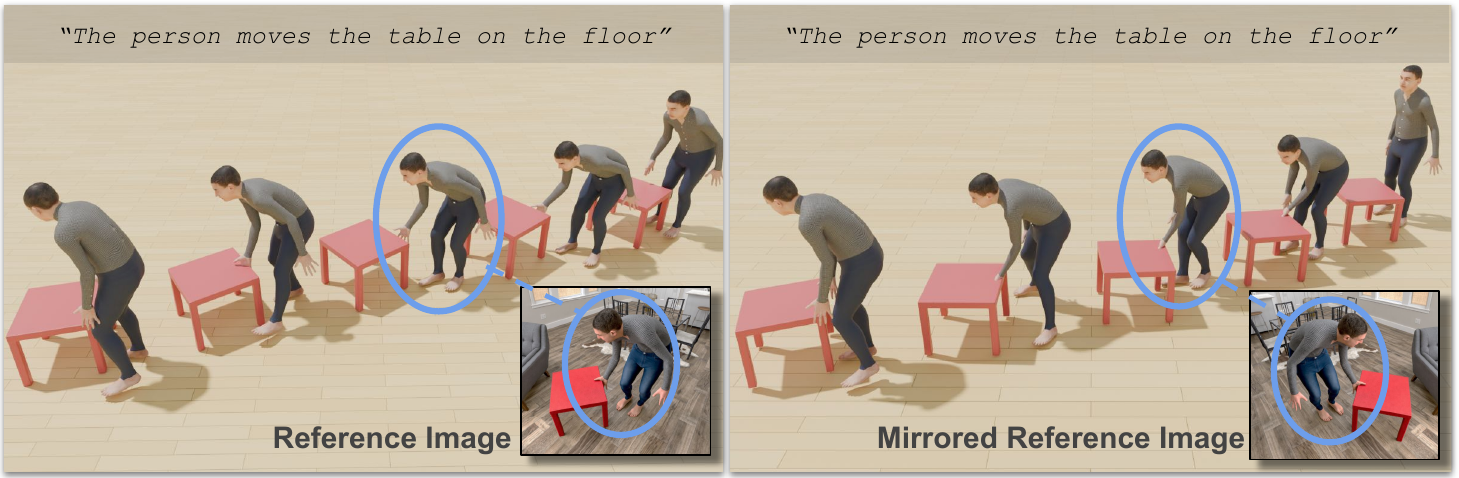}
\vspace{-2em}
\caption{
\textbf{Image-guided 4D HOI generation.}
Given a text prompt, object geometry, object waypoints, and a reference image, \method synthesizes a 4D human-object interaction sequence.
Text and waypoints specify the action and object trajectory, but leave fine-grained interaction details such as pose, contact, and layout ambiguous.
We resolve this ambiguity with a reference image that specifies the interaction snapshot.
To test whether \method follows this visual evidence, we keep the text prompt, object geometry, and waypoints fixed, and mirror only the reference image.
\method generates different motions that satisfy the corresponding snapshot: body pose, object pose, contact, and body-object layout change consistently with the mirrored reference.
This is achieved through spatio-temporal image conditioning, which separates spatial cues for the depicted interaction state from frame-aware cues for the surrounding motion.
Unlike single-token image conditioning, this preserves fine-grained visual evidence while generating the HOI sequence.
}
\Description{
Teaser showing image-conditioned human-object interaction generation.
For the same text prompt, object geometry, and sparse waypoints, the method generates different motions when given an original or mirrored reference image.
}
\label{fig:teaser}
\vspace{2.5em}
\end{teaserfigure}

\begin{abstract}
{\renewcommand{\thefootnote}{$\dagger$}%
\footnotetext{Work done during an internship at Meta, Zurich.}}%
\normalsize{
Generating human-object interactions (HOI) is central to character animation, robotics, AR/VR, and embodied AI. Controlled HOI generation has several practical implications and
recent methods synthesize motion from language, object geometry, and sparse geometric constraints, providing control over action semantics and object trajectories. 
However, these signals underspecify the interaction: the same prompt and trajectory can correspond to different grasps, approaching directions, body poses, object poses, contacts, and body-object layouts. 
We address this ambiguity by using a reference image as a visual specification of the desired interaction snapshot. 
However, a single global image representation conflates distinct interaction cues and conditions all motion frames on the same visual evidence.
Thus, we introduce \method, a diffusion-based HOI generator that decomposes the image conditioning signal \emph{spatio-temporally}.
For \emph{spatial conditioning}, \method extracts supervised interaction-state tokens that capture body pose, object pose, and body-object contact and spatial relationships at the depicted frame. 
For \emph{temporal conditioning}, it computes frame-aware tokens by querying image patches for each generated frame, allowing different parts of the sequence to attend to different visual cues from the same image. 
To balance image, text, and waypoint cues, \method uses role-aware conditioning: text, waypoints, and interaction-state tokens are conditioned through separate AdaLN streams, while frame-aware visual tokens are cross-attended with the learnable motion tokens.
Since existing HOI motion datasets lack paired images, we build a synthetic motion-to-image rendering pipeline from FullBodyManipulation (FBM) dataset sequences and introduce an image-adherence metric that evaluates whether generated motions match the reference
snapshot. 
Experiments on FBM and BEHAVE
show that \method improves fine-grained interaction control over single-token and uniformly image-conditioned baselines while preserving waypoint-following and motion quality. 
Code and models will be released at \textcolor{magenta}{\url{https://imagin4d.github.io}}.
}
\end{abstract}

\maketitle

\section{Introduction}
\label{sec:intro}

Generating 4D human-object interactions (HOI) is central to character animation, AR/VR, robotics, and embodied AI. 
This is challenging, because interactions need to look natural both spatially and temporally; 
that is, bodies and objects need to be posed with realistic contacts and proximal relationships in 3D space, and also move realistically over time. 
Moreover, it is challenging to computationally describe and exploit such spatial and temporal cues for effectively controlling the generation process.

Recent diffusion-based methods~\cite{xu2023interdiff,li2024chois} synthesize plausible HOI motion from text, object shape, and sparse waypoints, providing control over action semantics and object trajectories. 
However, these signals underspecify interactions; the same prompt and trajectory can correspond to different grasps, approaching directions, body poses, object poses, and contact or spatial configurations. 
These details matter for instruction-following agents, e.g., grasping a mug by its handle or opening a drawer correctly.

To reduce ambiguities one can manually specify 3D HOI keyposes. 
However, handcrafting these is cumbersome, and capturing them via MoCap is expensive. 
Instead, we use a reference image to intuitively represent a desired interaction ``snapshot;''
this may be a photo, a rendering, a sketch, or AI-generated. 
Thus, our goal is to synthesize HOI motion conditioned on a text prompt, object geometry, sparse waypoints, and a reference image of an HOI snapshot; see Fig.~\ref{fig:teaser}.

Similar conditioning has been explored by ViHOI~\cite{cai2026vihoi} concurrently to us.  
Specifically, ViHOI extracts a visual token from reference images to guide motion generation. 
However, this design has two limitations:
First, extracting a single global token abstracts distinct, fine-grained, spatial cues.
Second, these cues influence not only the timeframe of the reference snapshot, but also the motion frames before and after this, because motion is continuous and coherent. 
Consequently, as we show in our experiments, extracting a single global token compromises the fine-grained control necessary for motion generation.

Thus, we introduce \method, a diffusion-based HOI generator with \emph{spatio-temporal image conditioning}. 
Our key idea is that effective image conditioning needs to have both spatial and temporal influence on motion generation. 
For \emph{spatial influence}, \method extracts multiple, supervised spatial tokens that capture 
body pose, object pose, and body-object contact and spatial relationships.
To obtain these tokens, we introduce a ``Spatially-Factorized Image Encoder'' (SFIE) that applies separate Q-Former heads~\cite{li2023blip2} to image patches.
Therefore, the motion generator uses multiple, fine-grained tokens for the reference snapshot instead of just a single one. 
However, this informs only the snapshot timeframe.

For \emph{temporal influence} we need to inform all other frames as well. 
Thus, we introduce \emph{frame-aware tokens}, by querying image patches separately for each motion frame, conditioned on the frame index and text prompt. 
Unlike spatial tokens, frame-aware tokens are not supervised with explicit targets.
Instead, they are learned end-to-end through the denoising objective, allowing the model to learn image cues that are informative for each motion frame.

The spatial and frame-aware cues encode low-level details of interactions.
However, motion generators typically use high-level cues, i.e., text for action semantics and waypoints for the object trajectory. 
To prevent these cues from dominating over each other, \method does not concatenate conditioning tokens, but conditions on them differently, via a novel ``\emph{role-aware conditioning}''.
Specifically, spatial tokens, waypoints, and text use separate \emph{AdaLN} streams~\cite{peebles2023dit}, preserving their roles rather than mixing them.
In contrast, the frame-aware tokens are \emph{cross-attended} with the learnable motion tokens;
this selectively adapts the influence of frame-dependent cues on each motion token.

Training image-conditioned HOI synthesis %
requires images paired with motion data. 
However, existing HOI datasets contain MoCap data without paired images. 
We thus render images from FullBodyManipulation (FBM) sequences~\cite{li2023omomo} using body textures~\cite{black2023bedlam}, objects, and indoor scenes~\cite{straub2019replica}. 
For evaluation, standard metrics such as FID, R-precision, and contact F1 measure motion quality, semantic alignment, or contact accuracy, but do not measure the quality of image conditioning. 
Thus, we introduce the \emph{image-adherence metric} that evaluates to what extent the generated motion matches the reference snapshot.

We evaluate \method on the challenging FBM and BEHAVE \cite{bhatnagar2022behave}, with held-out object categories and cross-domain image inputs. 
Results show that %
the global image conditioning with a single token as in prior (concurrent) work improves 
motion metrics such as FID and R-Precision
but fails to control the depicted interaction state. 
In contrast, our spatially factorized tokens substantially improve image adherence, and adding frame-aware tokens further improves motion quality and contact accuracy.
As a downstream application, we retrain the image branch on line drawings to obtain a sketch-to-motion variant, showing that the same conditioning mechanism extends beyond RGB references to user-editable visual inputs.

In summary, our contributions are:
\begin{itemize}[leftmargin=*, topsep=0pt]
    \item 
    We introduce a HOI generator with a novel \emph{spatio-temporal image conditioning}, 
    using a reference interaction snapshot for controllable motion generation.
    \item 
    We introduce \emph{role-aware conditioning}: spatial tokens, waypoints, and text use separate AdaLN streams, and frame-aware tokens use cross-attention, improving adherence and motion quality.
    \item 
    We also introduce a synthetic \emph{motion-to-image rendering} pipeline and an \emph{image-adherence metric} for evaluating to what extent 
    the generated HOI motions match the reference interaction snapshot. 
\end{itemize}

\section{Related Work}
\label{sec:related}

\zheading{Data for dynamic HOI}
Whole-body HOI datasets~\cite{taheri2020grab,bhatnagar2022behave,huang2022intercap,jiang2023fullbody,li2023omomo,zhao2024imhd,lv2024himo,kim2025parahome,zhang2024hoim3,lu2025humoto} capture paired human-object motion, while hand-object benchmarks~\cite{fan2023arctic,liu2022hoi4d,zhan2024oakink2,liu2024taco} focus on manipulation and scene-aware corpora.  Some datasets \cite{hassan2019prox,wang2022humanise,jiang2024trumans} place motion in 3D environments. InterAct~\cite{xu2025interact} unifies several sources under a representation. 
These datasets support learning motion, contact, and scene constraints, but they are not designed for %
image-conditioned HOI generation; they do not provide motion-frame-aligned images that specify a target interaction state. 
We render conditioning images from FullBodyManipulation~\cite{li2023omomo} at known interaction frames using textured bodies~\cite{black2023bedlam}, objects, and indoor scenes~\cite{straub2019replica}.

\zheading{Controllable human motion generation}
Text-to-motion diffusion models~\cite{ho2020ddpm,tevet2023mdm,zhang2024motiondiffuse,chen2023mld} and tokenized generators~\cite{zhang2023t2mgpt,jiang2023motiongpt,guo2024momask,zhang2023remodiffuse,zhang2023finemogen} synthesize human motion from language. 
Additional control is introduced through guidance~\cite{karunratanakul2023gmd}, joint or keyframe constraints~\cite{xie2024omnicontrol,cohan2024condmdi}, sparse trackers~\cite{Barquero_2025_CVPR}, trajectory conditioning~\cite{shafir2024priormdm,wan2024tlcontrol,karunratanakul2024dno}, or masked control~\cite{tessler2024maskedmimic}. These interfaces work when the desired motion is expressible as body trajectories, masks, or key poses. They become cumbersome for HOI because the target state couples human pose, object pose, body-object layout, and contact. A reference image specifies this coupled state directly.

\zheading{4D Human-object interaction synthesis}
HOI synthesis models human motion, object motion, and their coupling through contact. Prior work studies hand-object manipulation~\cite{taheri2022goal,taheri2024grip,zhou2022toch,zhang2024graspxl,christen2024diffh2o}, whole-body object interaction~\cite{zhang2022couch,ghosh2023imos,li2023omomo,xu2023interdiff}, and language- or trajectory-conditioned 4D HOI generation~\cite{diller2024cghoi,peng2025hoidiff,cha2024text2hoi,song2024hoianimator,li2024chois,ron2025hoidini}. 
CHOIS~\cite{li2024chois} is the closest non-image-based baseline, since it controls long-horizon interactions with text, object geometry, and object waypoints. Other methods incorporate scene affordances, interaction fields, simulation, or priors distilled from image, video, and VLM models~\cite{wang2024affordmotion,jiang2024trumans,jiang2024lingo,yi2024tesmo,kulkarni2024nifty,yuan2023physdiff,wang2023physhoi,xu2025intermimic,li2024genzi,xu2024interdreamer,li2024zerohsi,zhang2025interactanything}. These methods improve realism and controllability, but most expose control through text, trajectories, contact maps, key poses, scene geometry, or priors. Our work instead uses a single reference image as the specification of the desired interaction state.

\zheading{Image-conditioned HOI and motion synthesis}
Image conditioning is used in 2D generation to constrain appearance, layout, identity, pose, or motion beyond text~\cite{ye2023ipadapter,mou2024t2iadapter,li2023gligen,huang2023composer,hu2024animateanyone,xu2024magicanimate,zhu2024champ}. Its use for 4D HOI generation is recent. ViHOI~\cite{cai2026vihoi} is closest to our setting: it extends a CHOIS-style generator with a frozen VLM and Q-Former adapter, then injects image-text tokens into the motion model. MP-HOI~\cite{wang2026mphoi} fuses text, image, and pose priors, whereas SIGHT~\cite{gavryushin2025sight} and IKMo~\cite{zhao2025ikmo}  use image and text conditioning to guide the motion. While these methods show that images can guide motion, they typically treat the image as a global condition, sparse anchor, or external prior. In HOI, this discards structure: the same image contains contact, body pose, object pose, body-object offset, and layout, and these cues matter at different times. Our method preserves this structure through supervised interaction-state tokens and frame-aware visual retrieval.

\section{Method}
\label{sec:method}

\begin{figure*}[t]
    \centering
    \includegraphics[width=\textwidth]{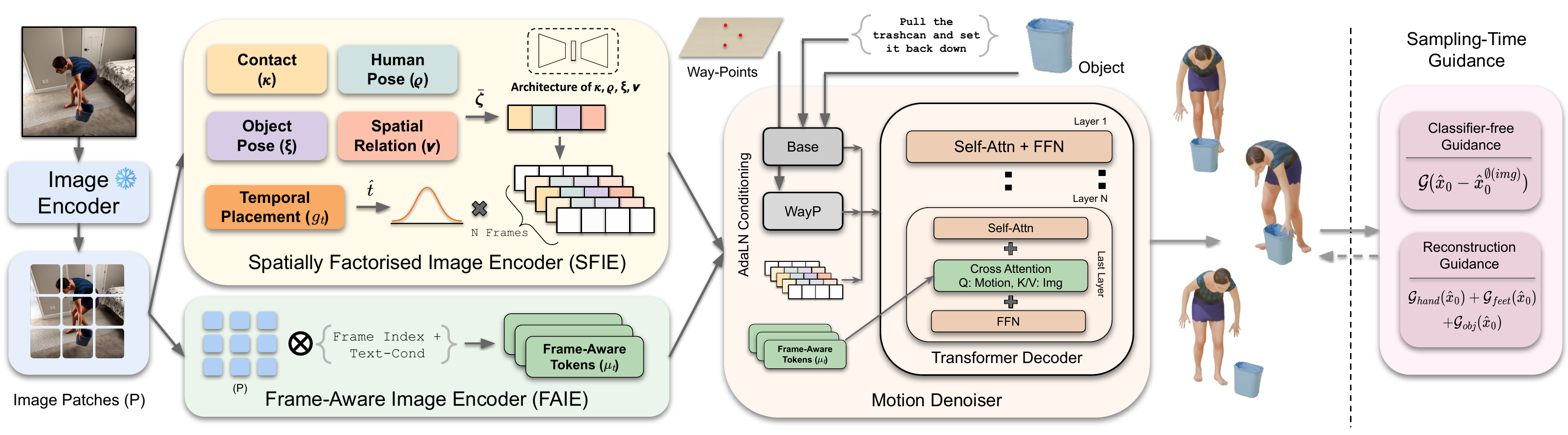}
    \caption{
        \textbf{\method overview (Sec.~\ref{sec:method}).}
        Given a reference image $\mathcal{I}$, text prompt $y$, object geometry $\mathcal{O}$, and sparse waypoints $\mathcal{W}$, \method generates a 4D human-object motion sequence. A frozen image encoder extracts patch tokens $\mathbf{P}$ from $\mathcal{I}$. The Spatially Factorized Image Encoder (SFIE) reads these patches with role-specific queries and produces supervised latent tokens for contact $\boldsymbol{\kappa}$, human pose $\boldsymbol{\rho}$, object pose $\boldsymbol{\xi}$, and body-object spatial relation $\boldsymbol{\nu}$. These tokens are trained to match role-autoencoder latents derived from the paired motion sequence. Their concatenated summary $\bar{\boldsymbol{\zeta}}$ predicts the reference frame $\hat{t}$ depicted by the image. In parallel, the Frame-Aware Image Encoder re-queries $\mathbf{P}$ with frame- and text-conditioned queries to produce per-frame visual tokens $\boldsymbol{\mu}_t$. The motion denoiser routes conditions by role: base conditioning, waypoint features, and window-gated spatial image evidence modulate transformer layers through separate AdaLN streams, while $\boldsymbol{\mu}_t$ enters through late cross-attention. Sampling-time guidance improves image adherence }%
    \label{fig:overview}
\end{figure*}

\method generates a human-object motion %
from a text prompt $\mathbf{Y}$, object shape $\mathcal{O}$, sparse trajectory waypoints %
$\mathcal{W}$, and a reference image $\mathcal{I}$.
The output is a $T$-frame %
motion sequence $\mathbf{x}_0\in\mathbb{R}^{T\times D}$,
where $D$ denotes the motion dimensionality of one frame.
The reference image depicts a %
desired interaction snapshot, corresponding to frame $t^\star$ in the target sequence.
During training, $t^\star$ is known from the paired image-motion data; at test time, it is predicted.

The reference image resolves interaction ambiguities left by text and sparse waypoints, including human pose, object pose, and body-object contact and layout.
\method uses two image representations.
First, a \emph{Spatially Factorized Image Encoder} (SFIE) extracts supervised role-specific spatial tokens for the reference snapshot.
Second, \emph{frame-aware %
tokens} re-query the same image patches for each motion frame.
The denoiser routes text, sparse waypoints, and window-gated spatial image evidence through separate AdaLN~\cite{peebles2023dit} streams, and injects frame-aware visual tokens through late cross-attention.
Fig.~\ref{fig:overview} summarizes the architecture.

\subsection{Conditional Motion Formulation}
\label{sec:prelim}

Let $\mathbf{x}_0\in\mathbb{R}^{T\times D}$ denote a %
HOI motion sequence.
Following CHOIS \cite{li2024chois}, sparse controls are represented by a binary mask $\mathbf{m}\in\{0,1\}^{T\times D}$, where $m_{t,d}=0$ denotes a input-specified value and $m_{t,d}=1$ denotes a value to synthesize.
The sparse conditioning tensor is $\mathbf{x}_{\mathrm{cond}}=(1-\mathbf{m})\odot\mathbf{x}_0$.
The sparse tensor and mask tell a diffusion denoiser which motion values are user-specified and which must be generated.
In our setting, $\mathbf{x}_{\mathrm{cond}}$ contains the initial human and object pose, sparse object waypoints, and final object translation. 
Object shape is encoded with a BPS encoder~\cite{prokudin2019bps}, yielding an object descriptor $\mathcal{O}_{BPS}$. 

We use conditional diffusion for motion generation.
At diffusion step $n$, the denoiser, $\mathcal{D}_{\theta}$, receives a noisy motion sequence $\mathbf{x}_n$ and predicts the clean motion,
$\hat{\mathbf{x}}_0=\mathcal{D}_{\theta}(\mathbf{x}_n,n,\mathbf{c})$,
where 
$\mathbf{c} = \{ \mathbf{Y}, \mathcal{O}_{BPS}, \mathbf{x}_{\mathrm{cond}} \}$ 
contains text, %
object shape, %
sparse controls. %

\subsection{Spatio-Temporal Image Conditioning}
\label{sec:image_conditioning}

\qheading{Spatially Factorized Image Encoder (SFIE)}
A pooled image embedding entangles pose, contact, and layout, which constrain the motion.
Thus, we encode the reference image into role-specific spatial tokens.
A frozen DINOv2 image encoder~\cite{oquab2023dinov2} extracts patch tokens $\mathbf{P}\in\mathbb{R}^{S\times C}$ from $\mathcal{I}$,
where $S$ denotes image patches and $C$ the feature dimension.
For each role $h\in\mathcal{H}$, a lightweight Q-Former~\cite{li2023blip2} reads $\mathbf{P}$ with learned queries $\mathbf{Q}_h$: \begin{equation} \mathbf{F}_h=\mathrm{QFormer}_h(\mathbf{P};\mathbf{Q}_h), \qquad h\in\mathcal{H}. \label{eq:image_factors} \end{equation} Here, $\mathbf{F}_h$ denotes the token output for role $h$. For human pose, object pose, contact, and layout, these outputs are $\boldsymbol{\rho}$, $\boldsymbol{\xi}$, $\boldsymbol{\kappa}$, and $\boldsymbol{\nu}$.
Each token is supervised to match the latent code of a lightweight role autoencoder trained on the corresponding ground-truth interaction quantity.
We concatenate and project the tokens into $\bar{\boldsymbol{\zeta}}$, used by the reference-frame localizer and image-AdaLN stream.

\zheading{Role supervision}
Each role is supervised at the ground-truth reference frame $t^\star$ using targets derived from the paired motion sequence.
The contacts decoders reconstruct 
    hand contact positions and binary contact flags.
The human-pose decoder %
    predicts main body-joint pose.
The object-pose decoder %
    predicts object translation and orientation.
The spatial-relation decoder %
    predicts per-joint unit vectors from body joints to the object center.
The SFIE loss is
\begin{equation}
\begin{aligned}
    \mathcal{L}_{\mathrm{SFIE}}
    &=
    \lambda_{\mathrm{con,pos}}\mathcal{L}^{\mathrm{mse}}_{\mathrm{con,pos}}
    +\lambda_{\mathrm{con,flag}}\mathcal{L}^{\mathrm{bce}}_{\mathrm{con,flag}}
    +\lambda_{\mathrm{hum}}\mathcal{L}^{\mathrm{mse}}_{\mathrm{hum}}
    \\
    &\quad
    +\lambda_{\mathrm{obj}}\mathcal{L}^{\mathrm{mse}}_{\mathrm{obj}}
    +\lambda_{\mathrm{spa}}\mathcal{L}^{\mathrm{cos}}_{\mathrm{spa}},
\end{aligned}
\label{eq:sfie_loss}
\end{equation}
where $\lambda$ denotes loss weights. The contact-position, contact-flag, human-pose, and object-pose losses supervise their corresponding decoder outputs, using MSE for continuous quantities and BCE for binary contact flags. The spatial loss $\mathcal{L}^{\mathrm{cos}}_{\mathrm{spa}}$ uses cosine distance between predicted and ground-truth joint-to-object unit vectors, making the spatial token encode body-object layout rather than absolute distance.

\qheading{Reference-frame localizer}
The reference image depicts one moment of the target sequence, but the corresponding frame is unknown at test time.
We predict it from the spatial-token summary:
\begin{equation}
    \boldsymbol{\pi} =
    \mathrm{softmax}\!\left(W_2\,\mathrm{GELU}(W_1\bar{\boldsymbol{\zeta}})\right)
    \in \mathbb{R}^T .
    \label{eq:fp}
\end{equation}
Here, $W_1,W_2$ are MLP weights, and $\pi_k$ is the $k$-th entry of $\boldsymbol{\pi}$, giving the probability that frame $k$ matches the reference image.
During training, we supervise this distribution with a Gaussian-smoothed target centered at $t^\star$:
\begin{equation}
    \mathcal{L}_{\mathrm{fp}}=-\sum_{k=0}^{T-1}q_k\log\pi_k, 
    \qquad
    q_k=\frac{\exp\left(-(k-t^\star)^2/(2\sigma_q^2)\right)}
    {\sum_{r=0}^{T-1}\exp\left(-(r-t^\star)^2/(2\sigma_q^2)\right)} .
    \label{eq:fp_loss}
\end{equation}
The predicted frame is $\hat{t}=\arg\max_k\pi_k$.
All image-window operations use $\hat{t}$ during both training and inference, so the denoiser is trained under the same localization errors it sees at test time.
Gradients from the denoising loss are stopped through $\hat{t}$; the localizer is trained only with Eq.~\eqref{eq:fp_loss}.

\qheading{Window-gated spatial image conditioning}
The spatial tokens describe the reference snapshot and constrain motion most strongly near $\hat{t}$.
Applying them uniformly can 
falsely constrain several frames before and after $\hat{t}$
toward the depicted pose and contact state.
We therefore gate the spatial-token summary with a temporal window:
\begin{equation}
  \tilde{\boldsymbol{\zeta}}_t
  =
  r_t\, w_{\sigma_g}(t-\hat{t})\,\bar{\boldsymbol{\zeta}} .
  \label{eq:gated_image_embedding}
\end{equation}
Here, $w_{\sigma_g}$ is a smooth unit-peak temporal window centered at $\hat{t}$.
The gate $r_t$ is set to $0$ at frames with specified waypoint constraints and to $1$ otherwise.
This prevents image modulation from competing with explicit trajectory constraints.

\qheading{Frame-Aware Image Encoder (FAIE)}
The supervised spatial tokens summarize the reference snapshot, but do not provide frame-specific %
evidence for the rest of the sequence.
Frames near $\hat{t}$ require precise contact and pose cues, while earlier or later frames rely more on object identity, coarse layout, or approach direction.
We therefore re-query image patches separately for each motion frame:
\begin{equation}
    \boldsymbol{\mu}_t=
    \mathrm{QFormer}_{\mathrm{fvt}}
    \left(\mathbf{P};\mathbf{Q}_{\mathrm{fvt}}(t,\mathbf{e}_y)\right), 
    \qquad
    \mathbf{M}=\{\boldsymbol{\mu}_t\}_{t=1}^{T}.
    \label{eq:frame_visual_tokens}
\end{equation}
Here, $\mathbf{e}_y$ is the text embedding, $\mathbf{Q}_{\mathrm{fvt}}(t,\mathbf{e}_y)$ denotes frame- and text-conditioned queries, and $\mathbf{M}$ is the 
set of frame-aware image tokens used for cross-attention.
The frame-aware token module is trained only through the denoising objective.

\subsection{Role-Aware Motion Denoiser}
\label{sec:role_aware_denoiser}

The denoiser receives conditioning signals with different roles.
Text provides sequence-level action semantics.
Sparse waypoint constraints provide trajectory control.
Window-gated spatial image evidence provides local constraints around the reference snapshot.
Frame-aware %
tokens provide frame-dependent image evidence.
A shared conditioning path can let dense image evidence perturb frames where waypoints already specify the object trajectory.
We therefore route each signal according to its role.

Let $\mathbf{c}_{\mathrm{text}}$ denote the text-conditioning vector combined with the diffusion-step embedding.
Let $\mathbf{z}^{\mathrm{wpt}}_t$ denote the sparse-constraint vector from $\mathbf{x}_{\mathrm{cond}}$ and $\mathbf{m}$ at frame $t$.
AdaLN converts each conditioning signal into per-layer modulation parameters for transformer.
At transformer layer $\ell$, the AdaLN parameters for frame $t$ are
\begin{equation}
    \boldsymbol{\eta}^{(\ell)}_t
    =
    \mathrm{AdaLN}^{(\ell)}_{\mathrm{text}}(\mathbf{c}_{\mathrm{text}})
    +
    \mathrm{AdaLN}^{(\ell)}_{\mathrm{wpt}}(\mathbf{z}^{\mathrm{wpt}}_t)
    +
    \mathrm{AdaLN}^{(\ell)}_{\mathrm{img}}(\tilde{\boldsymbol{\zeta}}_t).
    \label{eq:tristream}
\end{equation}
Here, $\boldsymbol{\eta}^{(\ell)}_t$ contains shift, scale, and residual-gate parameters used by the transformer block.
Separate streams let the model adjust image modulation without changing parameters that encode text semantics or waypoint constraints.
The image AdaLN stream is zero-initialized, so training starts from the text-and-waypoint model.

Frame-aware %
tokens are routed differently.
Instead of modulating every transformer layer, the denoiser reads visual memory $\mathbf{M}$ through cross-attention in the final decoder layer.
Each motion token accesses frame-dependent image evidence after self-attention integrates temporal context and sparse trajectory constraints.

\subsection{Synthetic Image Generation}
\label{sec:data_generation}

No existing dataset provides human-object motion \emph{paired} with images, which is required for image-guided motion generation.
The most related datasts are FullBodyManipulation (FBM)~\cite{li2023omomo} and BEHAVE~\cite{bhatnagar2022behave}, which only provide motions. 
Thus, we render images using meshes of these datasets 
(FBM has SMPL-X~\cite{pavlakos2019smplx} and BEHAVE has SMPL-H~\cite{romero2017mano} bodies), inserting the posed object, and assigning sequence-level body and object appearances.
We create three image domains. \emph{MeshImg} renders the body and object on a white background. \emph{SceneImg} places the same posed body-object pair in Replica indoor scenes \cite{straub2019replica} and applies BEDLAM body textures \cite{black2023bedlam}. \emph{EditImg} applies FLUX.2-dev \cite{flux2025} to SceneImg renders to produce more photorealistic test-time references.
EditImg is used only for evaluation. Since BEDLAM textures are baked to the SMPL-X UV layout, SceneImg is used only for OMOMO, which provides SMPL-X bodies~\cite{pavlakos2019smplx}. BEHAVE provides SMPL-H bodies~\cite{romero2017mano}, so we use MeshImg for BEHAVE training and evaluation. For FBM, SceneImg is used for the main experiments and MeshImg for ablations and cross-domain analysis.

For each 120-frame window, we render a uniform temporal grid and one contact-centered frame $t^\star$. We define $t^\star$ as the contact-weighted temporal centroid of the left- and right-hand contact flags; see \supmat for details.
At training time, one rendered image is sampled per sequence per epoch; at evaluation time, the contact-centered frame is the canonical reference for image adherence.

\subsection{Training and Sampling}
\label{sec:training_sampling}

\qheading{Denoising loss}
The denoiser predicts $\mathbf{x}_0$ directly.
Since image conditioning should matter most near the reference snapshot, we upweight reconstruction errors around the predicted reference frame:
\begin{equation}
    \mathcal{L}_{\mathrm{diff}}
    =
    \mathbb{E}\!\left[
    \frac{1}{T}\sum_{t=1}^{T}
    \left(
    1+\lambda_{\mathrm{cf}}
    \exp\!\left(-\frac{(t-\hat{t})^2}{2\sigma_{\mathrm{cf}}^2}\right)
    \right)
    \left\|
    \mathbf{x}_{0,t}-\hat{\mathbf{x}}_{0,t}
    \right\|_1
    \right].
    \label{eq:diffusion_loss}
\end{equation}
Both this loss and the image-conditioning gate use $\hat{t}$, matching training and inference behavior.

\qheading{Auxiliary losses}
Following prior HOI generation work~\cite{li2024chois}, we add a forward-kinematics loss $\mathcal{L}_{\mathrm{FK}}$, an object-point loss $\mathcal{L}_{\mathrm{objpts}}$, and a foot-contact loss $\mathcal{L}_{\mathrm{feet}}$.
We also use an image-consistency loss $\mathcal{L}_{\mathrm{img}}$ at the ground-truth reference frame $t^\star$.
This loss compares interaction quantities decoded from the generated frame with the corresponding reference-image targets, including human pose, object pose, and body-object contact and layout.
Its weight increases at lower-noise diffusion steps, where $\hat{\mathbf{x}}_0$ is reliable enough for frame-level geometric supervision.
The full objective is:
\begin{equation}
    \begin{aligned}
    \mathcal{L}
    &=
    \mathcal{L}_{\mathrm{diff}}
    +\mathcal{L}_{\mathrm{SFIE}}
    +\lambda_{\mathrm{fp}}\mathcal{L}_{\mathrm{fp}}
    +\lambda_{\mathrm{img}}\mathcal{L}_{\mathrm{img}}
    \\
    &\quad
    +\lambda_{\mathrm{FK}}\mathcal{L}_{\mathrm{FK}}
    +\lambda_{\mathrm{objpts}}\mathcal{L}_{\mathrm{objpts}}
    +\lambda_{\mathrm{feet}}\mathcal{L}_{\mathrm{feet}} .
    \end{aligned}
    \label{eq:total_loss}
\end{equation}
All loss weights,
optimizer settings, learning-rate schedule, data augmentation, and remaining hyperparameters are reported in \supmat.

\qheading{Image classifier-free guidance}
We use classifier-free guidance on image condition \cite{ho2022classifier}.
During training, with probability $p_{\mathrm{drop}}$, we replace all image inputs by their null form: spatial tokens $\boldsymbol{\zeta}$ and frame-aware tokens $\mathbf{M}$ are zeroed.
Text, object shape, and sparse controls remain unchanged.
At sampling, we run two denoiser forwards: one with full condition and one with dropped image condition.
Guided output is
\begin{equation}
    \hat{\mathbf{x}}_0^{\mathrm{cfg}}
    =
    \hat{\mathbf{x}}_0
    +
    (s_{\mathrm{img}}-1)
    \left(
    \hat{\mathbf{x}}_0-\hat{\mathbf{x}}_0^{\setminus\mathrm{img}}
    \right),
    \label{eq:cfg}
\end{equation}
where $\hat{\mathbf{x}}_0$ is the full-conditioning estimate, $\hat{\mathbf{x}}_0^{\setminus\mathrm{img}}$ the one with image inputs replaced by their null form, and $s_{\mathrm{img}}$ is the image-guidance scale.
Setting $s_{\mathrm{img}}=1$ recovers the standard conditional prediction.

\qheading{Mesh-level hand and foot guidance}
Following CHOIS~\cite{li2024chois}, we apply reconstruction guidance during the last $K=20$ denoising steps to reduce hand-object penetration and foot-floor artifacts.
At each guided step, we compute a mesh-level penalty on the predicted clean motion and take a variance-scaled gradient step before sampling $\mathbf{x}_{n-1}$.
The penalty combines a hand term, which penalizes hand-object penetration and attracts the hand to the object at frames predicted to be in contact, and a foot term, which penalizes support-foot deviation from the floor.
We restrict this guidance to the final denoising steps because the mesh penalties assume that $\hat{\mathbf{x}}_0$ is already close to a valid clean motion.

\section{Experiments}
\label{sec:experiments}

\newcolumntype{C}[1]{>{\centering\arraybackslash}p{#1}}

\begin{table*}[!t]
    \centering
    \small
    \setlength{\tabcolsep}{2.5pt}
    \renewcommand{\arraystretch}{1.2}
    \scalebox{0.95}{%
    \begin{tabular}{ll*{3}{C{0.9cm}}cccccccccc}
    \toprule
        \textbf{Method} 
        & \textbf{Img Enc} 
        & \multicolumn{3}{c}{\textbf{Image Adherence (cm)}\,$\downarrow$} 
        & \multicolumn{1}{c}{\textbf{FID}\,$\downarrow$}
        & \multicolumn{3}{c}{\textbf{R-Precision}\,$\uparrow$}
        & \multicolumn{3}{c}{\textbf{Contact}\,$\uparrow$}
        & \multicolumn{2}{c}{\textbf{Interaction}\,$\downarrow$}
        & \multicolumn{1}{c}{\textbf{WPErr}\,$\downarrow$} \\
        \cmidrule(lr){3-5}
        \cmidrule(lr){6-6}
        \cmidrule(lr){7-9}
        \cmidrule(lr){10-12}
        \cmidrule(lr){13-14}
        \cmidrule(lr){15-15}
        &
        &
        \textbf{$A_{\text{GT}}$}
        & \textbf{$A_{\text{W10}}$}
        & \textbf{$A_{\text{Any}}$}
        &
        & R@1
        & R@2
        & R@3
        & $C_{\text{prec}}$
        & $C_{\text{rec}}$
        & $C_{F_1}$
        & FootS
        & HandP
        &  \\
    \midrule
        InterDiff~\cite{xu2023interdiff} 
        & -- 
        & -- & -- & -- 
        & 20.80 
        & -- & -- & 0.08 
        & 0.63 & 0.28 & 0.33 
        & 0.42 & 0.55 
        & 72.72 \\
        
        MDM~\cite{tevet2023mdm} 
        & -- 
        & -- & -- & -- 
        & 6.16 
        & -- & -- & 0.51 
        & 0.72 & 0.47 & 0.53 
        & 0.48 & 0.66 
        & 19.42 \\
        
        CHOIS~\cite{li2024chois}
        & --
        & -- & -- & --
        & 0.69
        & 0.322 & 0.534 & 0.64
        & 0.80 & 0.64 & 0.67
        & \textbf{0.35} & 0.59
        & \textbf{2.87} \\
    \midrule
        CHOIS+Img
        & CLIP
        & 19.90 & 16.83 & 13.08
        & 0.47
        & 0.333 & 0.544 & 0.688
        & 0.796 & 0.625 & 0.655
        & 0.41 & 0.58
        & 3.31 \\

        CHOIS+Img
        & Qwen2.5-VL
        & 20.64 & 17.40 & 13.11
        & 0.75
        & 0.317 & 0.510 & 0.653
        & 0.798 & 0.534 & 0.640
        & 0.40 & 0.62
        & 3.93 \\
        
        ViHOI$^\star$~\cite{cai2026vihoi}
        & Qwen2.5-VL
        & 20.55 & 17.61 & 13.33
        & 0.71
        & 0.313 & 0.517 & 0.670
        & 0.800 & 0.611 & 0.648
        & 0.42 & 0.59
        & 3.43 \\

        CHOIS+Img
        & DINOv2
        & 19.61 & 16.48 & 13.23
        & 0.63
        & 0.330 & 0.543 & 0.687
        & 0.805 & 0.587 & 0.636
        & 0.45 & 0.60
        & 3.70 \\

        \textbf{Ours (Temporal)}
        & DINOv2
        & 19.07 & 16.23 & 12.56
        & 0.52
        & 0.340 & 0.565 & 0.714
        & 0.805 & 0.669 & 0.697
        & 0.39 & 0.62
        & 6.37 \\

        \textbf{Ours (Spatial)}
        & DINOv2
        & 10.11 & 8.91 & 7.95
        & 0.30
        & 0.352 & 0.569 & 0.716
        & 0.811 & 0.621 & 0.663
        & 0.44 & 0.56
        & 6.28 \\

        \textbf{Ours (Spatial+Temporal)}
        & DINOv2
        & \textbf{8.43} & \textbf{7.65} & \textbf{7.45}
        & \textbf{0.28}
        & \textbf{0.365} & \textbf{0.582} & \textbf{0.726}
        & \textbf{0.823} & \textbf{0.633} & \textbf{0.677}
        & 0.43 & \textbf{0.53}
        & 5.69 \\
    \bottomrule
    \end{tabular}}
    \caption{\textbf{Evaluation on the FullBodyManipulation dataset (Sec.~\ref{sec:main_comparison}).}
        All methods are evaluated on the FullBodyManipulation~\cite{li2023omomo} test set using text, object geometry, and sparse object waypoints as conditions; image-conditioned methods additionally use one rendered \emph{SceneImg} reference at the contact-centered frame. 
        We report image adherence at the ground-truth frame ($A_{\text{GT}}$), within a $\pm10$-frame window ($A_{\text{W10}}$), and over the full sequence ($A_{\text{Any}}$), along %
        with motion quality (FID), text-motion alignment (R-Precision), hand-object contact, interaction artifacts (foot sliding, hand penetration), and waypoint error. 
        The image-adherence score is not applicable %
        for image-free methods. 
        ViHOI$^\star$ denotes our single-image adaptation of ViHOI \cite{cai2026vihoi}.
    }
    \label{tab:sota_omomo}
    \vspace{-0.3 em}
\end{table*}

\begin{table}[!t]
    \centering
    \small
    \renewcommand{\arraystretch}{1.0}
    \resizebox{\columnwidth}{!}{%
    \begin{tabular}{l c c c c c c c c}
    \toprule
        \textbf{Method} 
        & A$_\text{GT}$$\downarrow$
        & A$_\text{W10}$$\downarrow$
        & A$_\text{Any}$$\downarrow$
        & FID$\downarrow$
        & $\mathbf{C_p}$$\uparrow$
        & $\mathbf{C_r}$$\uparrow$
        & $\mathbf{C_{F_1}}$$\uparrow$
        & WP$\downarrow$\\
    \midrule
        CHOIS~\cite{li2024chois}
        & -- & -- & -- & 1.39 & 0.526 & 0.351 & 0.375 & \textbf{3.05} \\
    \midrule
        CHOIS+Img
        & 22.02 & 18.81 & 13.71 & 1.57 & 0.521 & 0.361 & 0.386 & 4.07 \\
        ViHOI*~\cite{cai2026vihoi}
        & 22.97 & 20.34 & 14.55 & 1.44 & 0.518 & 0.348 & 0.378 & 4.49 \\
        \textbf{\method}
        & \textbf{12.40} & \textbf{12.23} & \textbf{11.18} & \textbf{0.76} & \textbf{0.538} & \textbf{0.381} & \textbf{0.397} & 4.93 \\
    \bottomrule
\end{tabular}%
}
\caption{\textbf{Comparison on BEHAVE (Sec.~\ref{sec:main_comparison}).}
    All image-conditioned methods are trained and tested on BEHAVE using single \emph{MeshImg} reference at the contact-centered frame. 
    We report image adherence at the ground-truth frame ($A_{\text{GT}}$), within a $\pm10$-frame window ($A_{\text{W10}}$), and over the full sequence ($A_{\text{Any}}$), plus FID, contact precision/recall/$F_1$, and waypoint error (WP). 
    ViHOI$^\star$ denotes our single-image adaptation of ViHOI. 
    \method improves adherence, FID, and contact over 
    CHOIS+Img and ViHOI$^\star$.
}
\label{tab:sota_behave}
\end{table}

\begin{table}[!t]
    \vspace{-2.0 em}
    \centering
    \small
    \renewcommand{\arraystretch}{1.0}
    \resizebox{\columnwidth}{!}{%
    \begin{tabular}{l c c c c c c c c}
    \toprule
        \textbf{Train $\rightarrow$ Test} 
        & A$_\text{GT}$$\downarrow$
        & A$_\text{W10}$$\downarrow$
        & A$_\text{Any}$$\downarrow$
        & FID$\downarrow$
        & $\mathbf{C_p}$$\uparrow$
        & $\mathbf{C_r}$$\uparrow$
        & $\mathbf{C_{F_1}}$$\uparrow$
        & WP$\downarrow$ \\
    \midrule
        MeshImg $\rightarrow$ MeshImg
        & 7.85 & 7.04 & 6.87 & 0.26 & 0.825 & 0.653 & 0.695 & 5.64 \\
        SceneImg $\rightarrow$ SceneImg
        & 8.43 & 7.65 & 7.45 & 0.28 & 0.823 & 0.633 & 0.677 & 5.69 \\
    \midrule
        MeshImg $\rightarrow$ SceneImg
        & 31.66 & 23.40 & 15.53 & 2.68 & 0.822 & 0.576 & 0.635 & 6.90 \\
        MeshImg $\rightarrow$ EditImg
        & 30.80 & 23.27 & 15.33 & 1.58 & 0.815 & 0.573 & 0.633 & 5.90 \\
        SceneImg $\rightarrow$ EditImg
        & 12.68 & 10.70 & 9.50 & 0.37 & 0.819 & 0.645 & 0.683 & 5.77 \\
    \bottomrule
    \end{tabular}%
    }
    \caption{\textbf{Cross-domain image generalization (Sec.~\ref{sec:cross_dataset}).}
        We train and test \method on FullBodyManipulation dataset across various 
        image-conditioning 
        domains while keeping motion data, text, object shape, and sparse waypoints fixed. 
        \emph{MeshImg} uses white-background body-object renders, \emph{SceneImg} adds indoor scenes, textures, and lighting, and \emph{EditImg} applies FLUX.2-dev image editing to SceneImg to increase photorealism. 
        Models trained with MeshImg transfer poorly to SceneImg and EditImg, whereas methods with SceneImg transfer much better to EditImg, supporting SceneImg for robust conditioning on photorealistic references.
}
\label{tab:crossdata}
\vspace{-2.0 em}
\end{table}

\qheading{Datasets \& Image protocol}
We follow the CHOIS protocol~\cite{li2024chois} on FullBodyManipulation (FBM)~\cite{li2023omomo} and BEHAVE~\cite{bhatnagar2022behave, xu2025interact}. 
FBM contains human-object MoCap sequences with nine training and four held-out object categories; we use the same train/test split, and text prompts as CHOIS. 
BEHAVE provides RGB-D human-object sequences, which we convert to the same 120-frame windowing convention.
Since these datasets do not provide reference images for our image-conditioned generation protocol, we use the rendered conditioning images described in Sec.~\ref{sec:data_generation}.
For the main FBM comparison and ablations, both training and testing use \textit{SceneImg}. \textit{MeshImg} and \textit{EditImg} are used for cross-domain image analysis; see Fig.~\ref{fig:dataset_render}. Applying FLUX on MeshImg often changes pose or object layout, so we generate EditImg from SceneImg; see \supmat.
BEHAVE is trained and evaluated with \textit{MeshImg} because BEHAVE uses SMPL-H bodies, which are incompatible with the SMPL-X UV layout used by BEDLAM textures. Rendering details and reference-frame selection are given in Sec.~\ref{sec:data_generation} and \supmat.

\qheading{Metrics}
We report image adherence, motion quality, text alignment, contact, interaction artifacts, and waypoint following. 
Image adherence is a new metric that we define to measure whether the generated sequence realizes the reference interaction state; 
we report $A_{\text{GT}}$ at $t^\star$, $A_{\text{W10}}$ within a $\pm10$-frame window, and $A_{\text{Any}}$ over the full sequence, all in cm. Motion quality and text alignment are measured by FID and R-Precision using the frozen CHOIS evaluator. Contact is measured by hand-object contact precision, recall, and $F_1$. Interaction artifacts are measured by foot sliding and hand-object penetration. Waypoint error measures object-trajectory error at sparse waypoint frames. For formal definitions, see \supmat.

\qheading{Baselines and implementation}
We compare against image-free HOI baselines: CHOIS~\cite{li2024chois}, InterDiff~\cite{xu2023interdiff}, and MDM~\cite{tevet2023mdm}. For image-conditioned baselines, we retrain CHOIS with a single pooled image token from CLIP~\cite{radford2021clip}, DINOv2~\cite{oquab2023dinov2}, and Qwen-VL~\cite{cai2026vihoi}. We also re-implement ViHOI~\cite{cai2026vihoi} (since code is not yet public), adapting its three-image input to our single-image setting and marking this row as ViHOI$^\star$; see \supmat\~ for details. 

\subsection{Main Comparison on FBM and BEHAVE}
\label{sec:main_comparison}

\begin{figure*}[t]
\centering
\includegraphics[width=\textwidth]{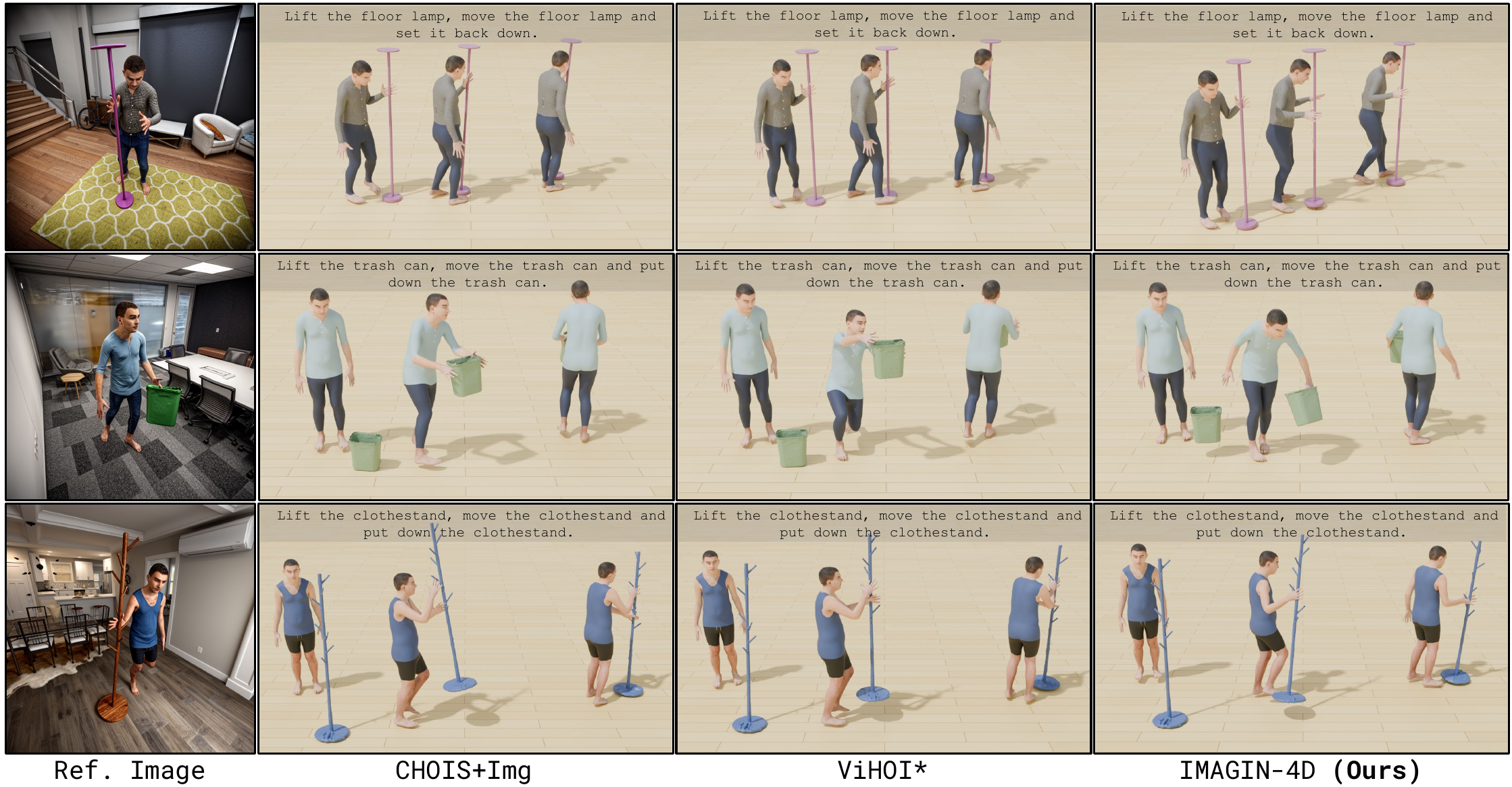}
\vspace{-2.0em}
    \caption{
    \textbf{Qualitative comparison on FullBodyManipulation (Sec.~\ref{sec:main_comparison}).}
    Each row shows a reference image (\emph{EditImage}) and generated motion from CHOIS+Img, ViHOI$^\star$, and \method, all conditioned on the same text prompt, object shape, waypoints, and reference image.
    Single-token image conditioning often misses fine-grained details, such as contact, %
    hand placement, or body-object layout, while \method better matches the depicted interaction state.
    }
    \Description{
    Qualitative comparison of image-conditioned human-object interaction generation. Each row contains a reference image and generated results from three methods for interactions with a floor lamp, trash can, and clothes stand. The proposed method better preserves the depicted contact, object pose, and body-object spatial layout.
    }
\label{fig:sota_comp_editimg}
\vspace{+1.5 em}
\end{figure*}

Table~\ref{tab:sota_omomo} evaluates whether image conditioning improves fine-grained interaction control on FBM without degrading motion quality or waypoint following. CHOIS remains the strongest waypoint follower, reaching $\text{WPErr}=2.87$\,cm because it has no image condition ``competing'' with other condition signals.
Image-free methods do not receive a reference image, so image-adherence metrics are reported only for image-conditioned methods.

Baselines using a single global image token, namely CHOIS+Img and ViHOI$^\star$, improve some global motion metrics such as FID but provide limited image adherence.
For CHOIS+Img and ViHOI$^\star$, $A_{\text{Any}}$ remains between $13.08$ and $13.33$\,cm across CLIP, DINOv2, and Qwen2.5-VL image encoders. ViHOI$^\star$ improves FID relative to CHOIS+Img with the same Qwen2.5-VL encoder, but its adherence remains comparable.
This indicates a representation bottleneck: a single image token compresses spatial evidence that is needed to recover body pose, object pose, and body-object contact and layout.

Our spatial encoder addresses this bottleneck with supervised interaction tokens localized at the conditioning frame. \textbf{Ours (Spatial)} reduces $A_{\text{Any}}$ from $12.56$ to $7.95$\,cm relative to \textbf{Ours (Temporal)} that uses global image representation from DINOv2 and improves FID from $0.52$ to $0.30$. Spatial tokens give strong reference-frame control, but they are fixed across the sequence. \textbf{Ours (Spatial+Temporal)} adds frame-aware tokens that query the image patches for each motion frame, testing whether image evidence can also improve the surrounding motion. Waypoint error increases relative to CHOIS because image and waypoint conditions compete for control; 
\supmat\~ analyzes this tradeoff.

Figure~\ref{fig:sota_comp_editimg} shows the qualitative failure mode under photorealistic EditImg conditioning. 
Baselines using global image representation with a single token often preserve action category but miss the depicted contact, %
object orientation, or hand-object layout. 

Table~\ref{tab:sota_behave} repeats the comparison on BEHAVE using MeshImg, testing whether 
spatial image factorization
transfers to a different motion distribution.
The same trend holds: spatial factorization improves image adherence over global image conditioning, showing that the gain is not specific to FBM. 
This isolates the effect of the image representation under the BEHAVE rendering protocol.

\subsection{Cross-Domain Image Generalization}
\label{sec:cross_dataset}

\begin{figure*}[p]
\centering
\includegraphics[width=\textwidth]{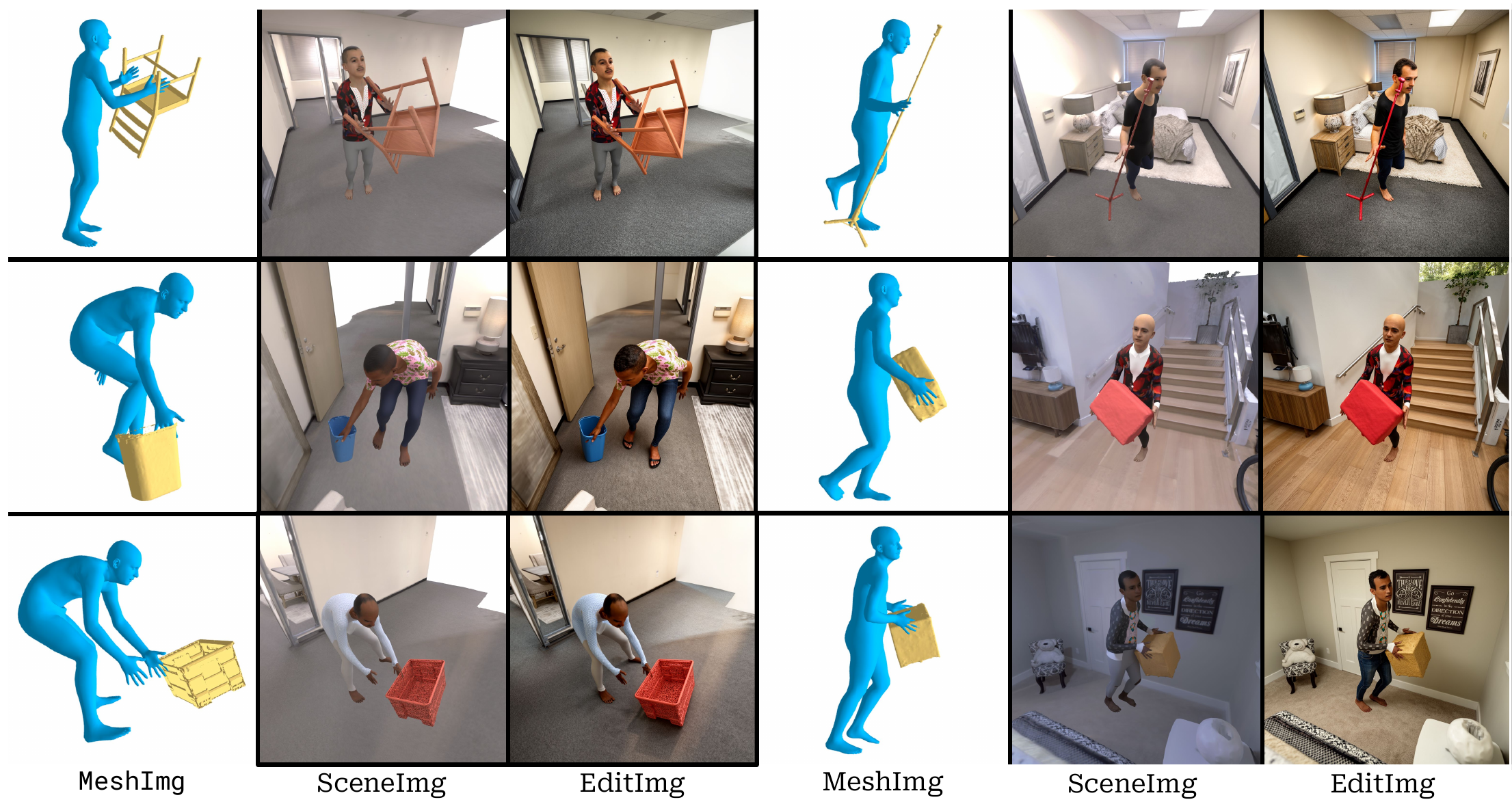}
\vspace{-2em}
\caption{
\textbf{Conditioning image domains (Sec.~\ref{sec:cross_dataset}).}
We render conditioning images from each ground-truth sequence from the FullBodyManipulation dataset and use the contact-centered frame for evaluation.
\emph{MeshImg} is a clean body-object render, \emph{SceneImg} adds Replica scenes, body textures, and posed objects, and \emph{EditImg} applies image editing for more photorealistic references.
SceneImg is used for evaluation, while MeshImg and EditImg analyze image-domain transfer.
}
\Description{}
\label{fig:dataset_render}
\end{figure*}

Table~\ref{tab:crossdata} tests whether image conditioning transfers across rendering domains. The motion ground truth is fixed; only the conditioning image changes. Training on MeshImg transfers poorly to SceneImg or EditImg, with $A_{\text{GT}}$ degrading by roughly $23$--$24$\,cm and FID increasing sharply. Training on SceneImg transfers much better to EditImg: $A_{\text{GT}}$ increases by only $4.25$\,cm relative to same-domain SceneImg. Together with Fig.~\ref{fig:dataset_render}, this supports using SceneImg as the main training domain: it contains enough scene, lighting, texture, and material variation to transfer to more photographic references. Detailed cross-domain analysis is provided in \supmat.

\subsection{Qualitative Results}
\label{sec:qualitative}

\begin{figure*}[p]
\centering
\includegraphics[width=0.86\textwidth]{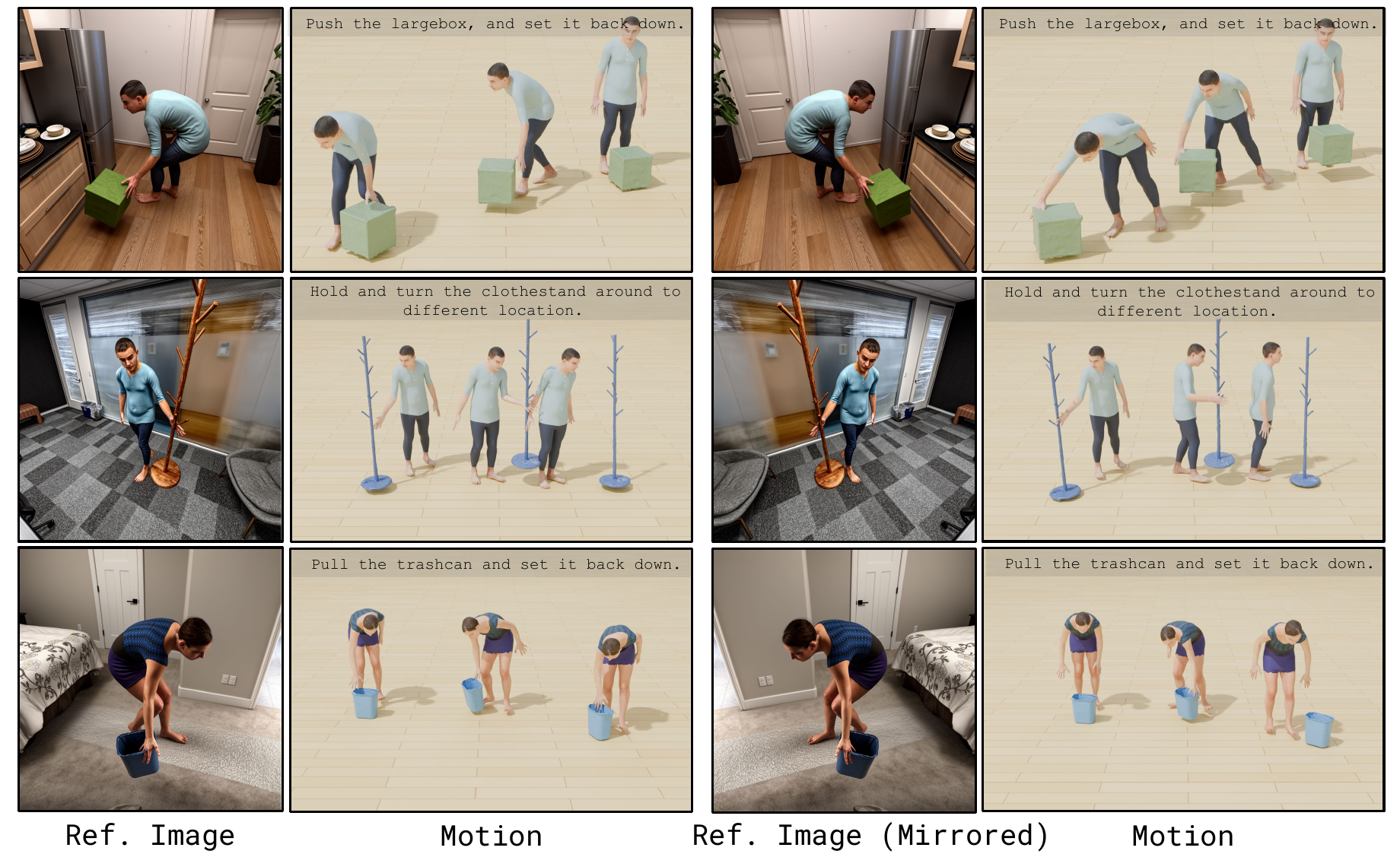}
\vspace{-1em}
\caption{
\textbf{Flip-image consistency test (Sec.\ref{sec:qualitative}).}
We horizontally flip only the reference image at inference time while keeping the text prompt, object geometry, and waypoints fixed.
The generated contact side and body-object layout change with the flipped image, showing that \method uses visual evidence rather than ignoring the image condition.
The motion is not an exact mirror because the unchanged non-image conditions must still be satisfied.
}
\Description{}
\label{fig:flip_motion}
\end{figure*}

Figure~\ref{fig:flip_motion} tests whether the generated motion causally depends on the image. We horizontally flip only the conditioning image at inference time, keeping text, waypoints, object geometry, conditioning frame, and random seed fixed. The generated grasp side and contact region change, showing that the model actually uses the image. The generated motion %
is not fully mirrored, %
because the non-image conditions remain unchanged and still need to be met; this is the desired behavior for a conditional generator that must reconcile image evidence with text, object geometry, and waypoints.

\begin{figure*}[p]
\centering
\includegraphics[width=0.86\textwidth]{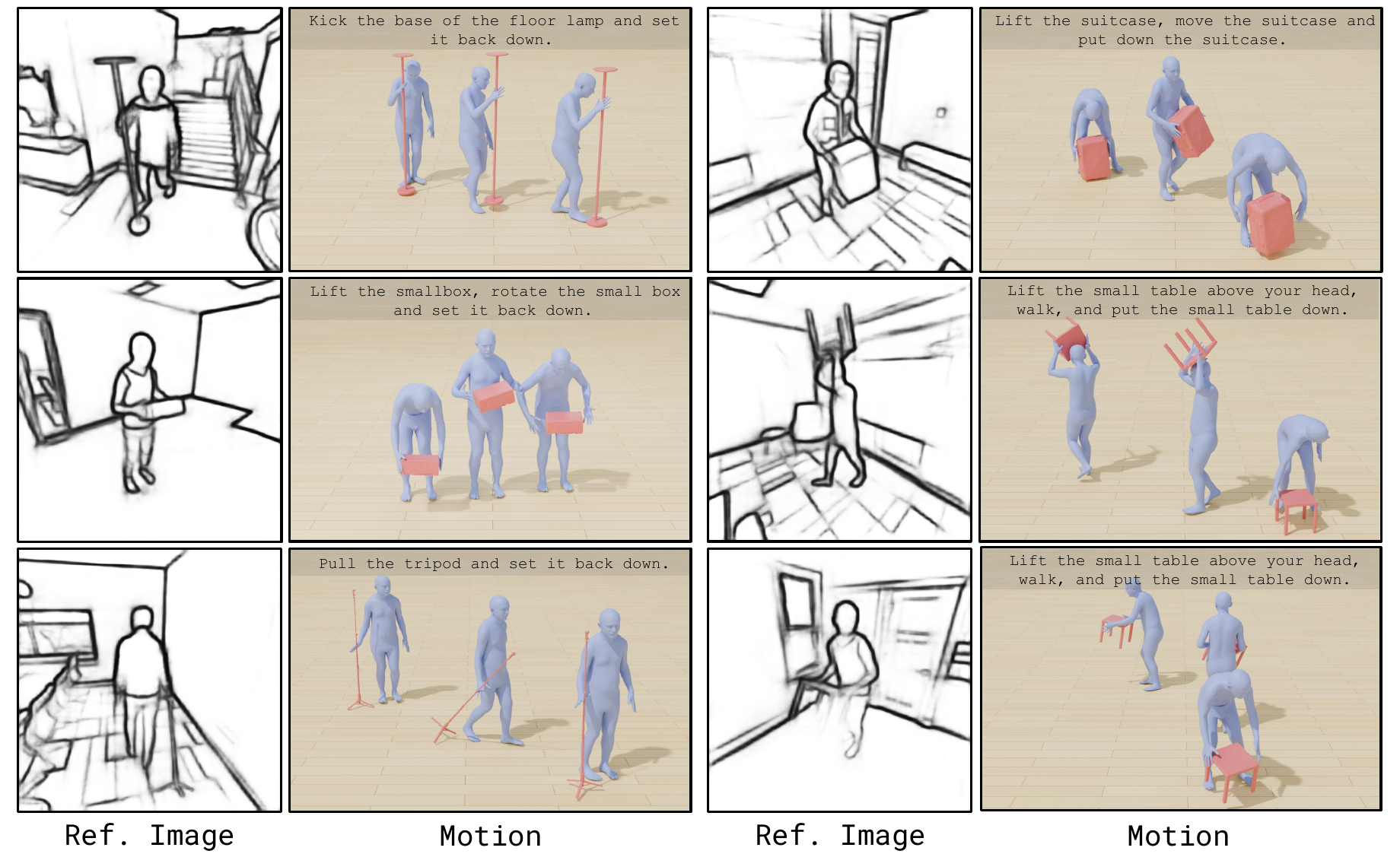}
\vspace{-1em}
\caption{
\textbf{Sketch-to-motion (Sec.\ref{sec:qualitative}).}
We replace the RGB reference image with a line drawing and retrain the model.
Despite removing texture, color, and scene appearance, the model preserves the depicted interaction layout and generates a complete motion sequence.
This shows that \method can also support sketch-based conditioning, where users specify the desired contact and body-object arrangement with a simple drawing.
}
\Description{}
\label{fig:sketch_to_motion}
\end{figure*}

Finally, Fig.~\ref{fig:sketch_to_motion} shows sketch-to-motion as a downstream authoring application. We retrain the image branch using line drawings from the same rendered reference frames. 
Image adherence is lower for sketch conditioning than for RGB conditioning because line drawings remove appearance cues that help localize contact and pose; quantitative results are provided in \supmat.
This shows that the method is not limited to photometric RGB inputs, although sketches remove appearance cues useful for contact and pose localization. 

\section{Conclusion}
\label{sec:conclusion}

We present \method, an image-guided HOI generator that uses a reference interaction snapshot to control 4D human-object motion. 
Our results show that reference images are useful for HOI generation only when their spatial cues are preserved and propagated across time.
A single pooled visual token can improve motion metrics like FID and R-Precision, but it collapses the fine-grained cues needed to recover the depicted contact, object pose, and body-object layout.
In contrast, spatially factorized tokens recover the reference interaction state, and frame-aware tokens extend this visual influence to the surrounding motion frames.
Role-aware conditioning then preserves these visual cues by routing image, text, and waypoint signals through separate streams rather than merging them into one conditioning signal.
Our rendering pipeline and image-adherence metric enable controlled evaluation across image domains, including sketches, and across datasets like FBM and BEHAVE.
Together, these results establish spatio-temporal image conditioning as a practical way to specify fine-grained human-object interactions beyond text and sparse waypoints, without sacrificing motion quality.
Code and trained models will be publicly released.

\clearpage

\clearpage

\begin{figure*}[p]
\centering
\includegraphics[width=\textwidth]{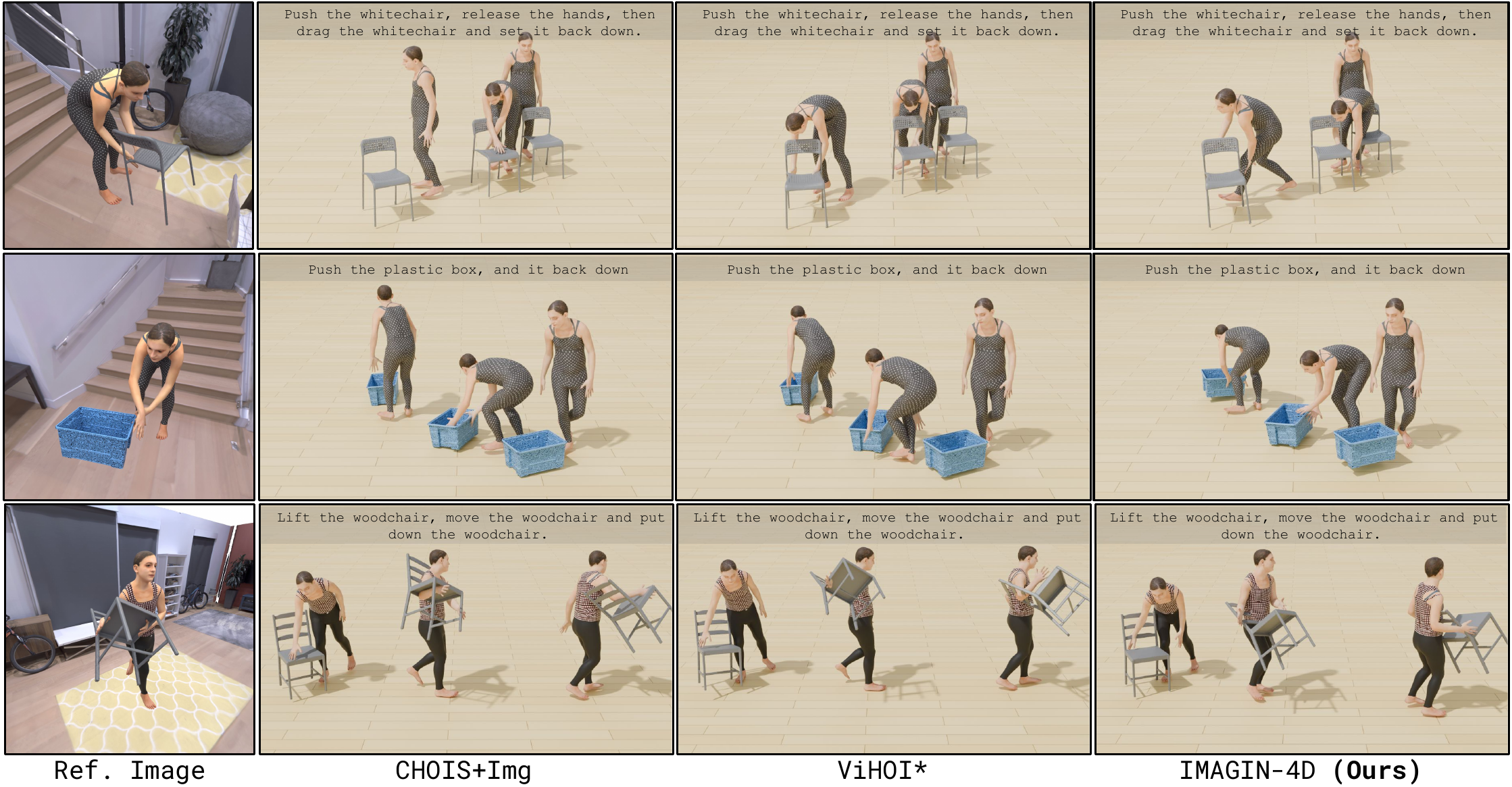}
\vspace{-2em}
\caption{
    \textbf{Qualitative comparison on FullBodyManipulation (SceneImg) (Sec.~\ref{sec:main_comparison}).}
    Each row shows the \emph{SceneImg} reference and generated motion from CHOIS+Img, ViHOI$^\star$, and \method under the same text prompt, object shape, waypoints, and reference image.
    Single-token image conditioning often misses contact, %
    hand placement, object orientation, or body-object layout.
    \method better matches the depicted interaction state.
}
\Description{}
\label{fig:sota_comp_sceneimg}
\end{figure*}

\clearpage

\noindent
{\small
\textbf{Acknowledgments:} We thank Ilya Petrov, Nikos Athanasiou, Angela Yao, Javier Romero, Soubhik Sanyal, and Robin Kips for insightful discussions and feedback. We also thank Joachim Tesch for help with BEDLAM textures. DT was supported by the ERC Starting Grant (project \mbox{STRIPES}, \mbox{101165317}).}

{
\balance
\bibliographystyle{ACM-Reference-Format}
\bibliography{arXiv}
}

\clearpage
\twocolumn[%
  \begin{center}
    {\Large\bfseries Supplementary Material\par}
    \vspace{6pt}
    {\Large \textsc{IMAGIN-4D}: Image-Guided Controllable Interaction Generation \par}
    \vspace{10pt}
  \end{center}
]
\noindent\textbf{Overview.}\quad
This supplement provides metric definitions, image-adherence details, data generation, baseline construction, architecture and training details, ablations, cross-domain analysis, BEHAVE protocol notes, sketch-conditioning results, qualitative probes, and failure cases.

\appendix

\section{Metrics}
\label{sec:supp_metrics}

We report five metric groups. Distances are reported in centimeters unless stated otherwise.

\qheading{Image adherence}
Image adherence measures whether the generated sequence realizes the interaction state depicted in the reference image. At the annotated conditioning frame $t^\star$, the reference image defines a target human-object configuration. We report three variants: $A_{\text{GT}}$, evaluated exactly at $t^\star$; $A_{\text{W10}}$, the best score within a $\pm10$-frame window around $t^\star$; and $A_{\text{Any}}$, the best score over the full 120-frame sequence. Lower is better. For details, see Sec.~\ref{sec:supp_image_metric}.

\qheading{Motion quality and text alignment}
FID and R-Precision are computed using the frozen CHOIS text-to-motion evaluator~\cite{li2024chois}. This keeps the protocol directly comparable to prior text-and-waypoint HOI generation methods.

\qheading{Contact}
Contact is measured by precision, recall, and $F_1$ between generated left/right hand contact flags and the ground-truth BEHAVE contact annotations.

\qheading{Interaction artifacts}
We report foot sliding and hand-object penetration. Foot sliding measures the motion of foot joints while they are predicted to be in floor contact. Hand-object penetration measures mesh-level hand penetration into the object using the rest-frame object SDF. For BEHAVE categories without rest-frame SDFs, hand penetration is not reported.

\qheading{Waypoint following}
Waypoint error measures the planar distance between the generated object trajectory and the prescribed sparse waypoints at waypoint timestamps. Since all controllable baselines receive waypoints, this metric measures whether image conditioning degrades trajectory control.

\section{Image Adherence Metric}
\label{sec:supp_image_metric}

Let the generated sequence be $\{(\mathbf{j}_t,\mathbf{o}_t)\}_{t=1}^{T}$, where $\mathbf{j}_t\in\mathbb{R}^{24\times 3}$ are body-joint positions and $\mathbf{o}_t\in\mathbb{R}^{3}$ is the object centroid. The reference image corresponds to the ground-truth interaction state $(\mathbf{j}^{\star},\mathbf{o}^{\star})$ at frame $t^\star$. Predicted and ground-truth poses are mapped to the same canonical interaction frame before evaluation.

For each generated frame $t$, we compute
\begin{equation}
d(t) =
0.8\, d_{\text{hum}}(t)
+
0.2\, \|\mathbf{o}_t-\mathbf{o}^{\star}\|_2 ,
\label{eq:adh_combined}
\end{equation}
where
\begin{equation}
d_{\text{hum}}(t)
=
\sum_{k=1}^{24}
w_k
\|\mathbf{j}_{t,k}-\mathbf{j}^{\star}_{k}\|_2 ,
\qquad
w_k =
\frac{
\exp\!\left(-\|\mathbf{j}^{\star}_{k}-\mathbf{o}^{\star}\|_2/\sigma\right)
}{
\sum_{r=1}^{24}
\exp\!\left(-\|\mathbf{j}^{\star}_{r}-\mathbf{o}^{\star}\|_2/\sigma\right)
},
\label{eq:adh_weights}
\end{equation}
with $\sigma=0.3$m. The weights emphasize joints close to the object in the reference interaction state, making the metric sensitive to contact-relevant body parts while still accounting for full-body configuration. We report
\begin{align}
A_{\text{GT}} &= d(t^\star), \\
A_{\text{W10}} &= \min_{|t-t^\star|\leq 10} d(t), \\
A_{\text{Any}} &= \min_{1\leq t\leq T} d(t).
\end{align}

$A_{\text{GT}}$ measures exact temporal alignment. $A_{\text{W10}}$ allows small timing shifts while still requiring the interaction to occur near the intended frame. $A_{\text{Any}}$ diagnoses whether the desired interaction is realized somewhere in the sequence even when temporal placement is imperfect.

\section{Data Generation Details}
\label{sec:supp_data}

FullBodyManipulation (FBM)~\cite{li2023omomo} and BEHAVE~\cite{bhatnagar2022behave} provide human-object motion but not reference images for image-conditioned generation. We therefore render reference images from ground-truth motion sequences.

\qheading{Reference-frame selection}
For each 120-frame window, we render frames from a uniform temporal grid and one contact-centered frame. The contact-centered frame is the temporal centroid of left- and right-hand contact flags:
\begin{equation}
t^\star =
\mathrm{round}
\left[
\frac{\sum_t t(c^{\mathrm{L}}_t+c^{\mathrm{R}}_t)}
{\sum_t (c^{\mathrm{L}}_t+c^{\mathrm{R}}_t)}
\right].
\end{equation}
If no hand contact occurs, we use the window center.

\qheading{Rendering domains}
We use three image domains. \emph{MeshImg} renders the posed body and object on a white background. \emph{SceneImg} places the same human-object pair in Replica indoor scenes~\cite{straub2019replica} with BEDLAM body textures~\cite{black2023bedlam}, object materials, and Blender Cycles shading. \emph{EditImg} applies a directive-prompt FLUX.2-dev~\cite{flux2025} image-editing pass to SceneImg, with prompts designed to preserve body pose, hand placement, object pose, and contact while changing appearance.

SceneImg is used for the main FBM training and evaluation because it contains scene context, lighting, shadows, clothing texture, and object appearance. MeshImg is used for controlled ablations and for BEHAVE, where SMPL-H bodies do not share the SMPL-X UV layout used by BEDLAM textures. EditImg is used only at test time to evaluate robustness to more photorealistic references.

\qheading{Scene placement and camera}
For SceneImg, we sample valid navigable scene locations, place the body-object pair on the local floor, and select a camera that maximizes human-object visibility while avoiding severe scene occlusion. Images are rendered at $512\times512$ using Blender Cycles. The Cycles render is deterministic for a fixed sequence, scene, and camera, so all methods receive identical conditioning images.

\begin{figure}[t]
\centering
\includegraphics[width=\columnwidth]{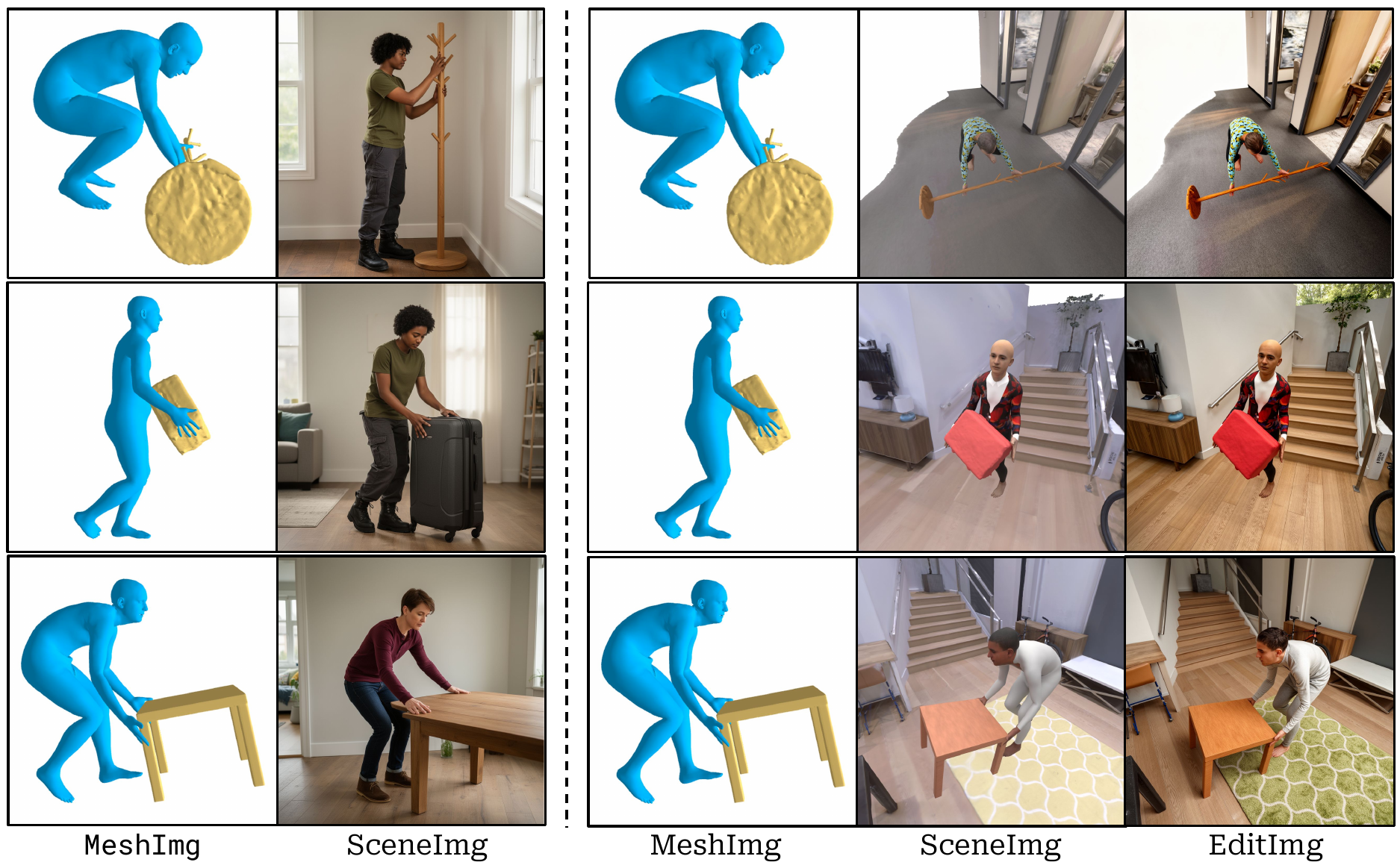}
\vspace{-1em}
\caption{\textbf{Why EditImg is derived from SceneImg.}
Columns 1--2 show direct MeshImg$\rightarrow$EditImg results. Columns 3--5 show the same interaction as MeshImg, SceneImg, and SceneImg-derived EditImg. Direct MeshImg editing can preserve the interaction in some cases, but can also change human pose, object pose, contact, or body-object layout. Since image adherence assumes that the reference preserves the target interaction geometry, we use the more reliable SceneImg$\rightarrow$EditImg protocol for evaluation.}
\Description{Single-column comparison of MeshImg editing and SceneImg editing. Each row shows a source interaction. The first two columns show MeshImg input and directly edited output. The last three columns show MeshImg, SceneImg, and SceneImg-derived EditImg. Some direct MeshImg edits change the interaction geometry, while another preserves it more faithfully.}
\label{fig:mesh_flux_failure}
\end{figure}

\qheading{Main-image protocol}
Unless otherwise stated, the FBM test image is SceneImg sampled at $t^\star$. Main FBM comparisons and architecture ablations are trained and evaluated on SceneImg. MeshImg and EditImg are reserved for cross-domain probes so that the domain-shift experiment changes the image distribution rather than the motion sequence.

\qheading{Why EditImg is derived from SceneImg}
We apply FLUX.2-dev editing to SceneImg rather than MeshImg to keep the evaluation geometry controlled. MeshImg editing sometimes produces plausible photographic images, but the edit can also change human pose, object pose, hand-object contact, or body-object layout. This breaks the image-adherence protocol, which assumes that the reference image preserves the target interaction geometry from the ground-truth motion. SceneImg already contains indoor context, shadows, body texture, and object appearance, so editing mainly changes visual realism while better preserving the interaction state. We therefore use SceneImg$\rightarrow$EditImg as the photorealistic transfer setting. Fig.~\ref{fig:mesh_flux_failure} shows two failure cases and one partially successful MeshImg edit.

\section{Baseline Construction}
\label{sec:supp_baselines}

\qheading{Image-free baselines}
We report the published FBM numbers from CHOIS~\cite{li2024chois}. These include CHOIS, MDM~\cite{tevet2023mdm}, and InterDiff~\cite{xu2023interdiff}, with MDM and InterDiff retrained for the text-and-waypoint HOI generation task by CHOIS.

\qheading{Pooled-image CHOIS baselines}
To test whether image conditioning helps under a simple interface, we retrain CHOIS with one pooled visual token. We use three encoders: CLIP~\cite{radford2021clip}, DINOv2~\cite{oquab2023dinov2}, and Qwen-VL~\cite{cai2026vihoi}. These rows isolate the effect of encoder choice when the conditioning interface is fixed to a single pooled token.

\qheading{ViHOI re-implementation}
The closest image-conditioned competitor is ViHOI~\cite{cai2026vihoi}, which compresses reference images into a single visual token using a Qwen-VL backbone and Q-Former adapter. We implement ViHOI following its published formulation and adapt its three-image input to our single-image setting for a matched comparison. We mark this adapted row as ViHOI$^\star$.

\qheading{Controlled encoder-interface comparison}
The Qwen-VL pooled baseline and ViHOI$^\star$ hold the visual backbone family fixed while changing the image interface from pooling to a Q-Former token. The CLIP and DINOv2 pooled baselines hold the single-token interface fixed while changing the encoder. These controlled rows separate encoder strength from the representational bottleneck of single-token conditioning.

\section{Training and Optimization Details}
\label{sec:supp_training_details}

\qheading{Backbone}
The denoiser is a 4-layer transformer with width $512$ and $4$ attention heads, trained as an $x_0$-prediction DDPM on 120-frame motion windows. Each frame contains object translation, object rotation, body joint positions, body joint rotations, and semantic contact channels. Object geometry is represented with a 1024-point BPS encoding~\cite{prokudin2019bps}.

\qheading{Conditioning}
The model uses three AdaLN streams: a base stream for diffusion step and text conditioning, a waypoint stream for sparse trajectory constraints, and an image stream for window-gated spatial image tokens. The waypoint and image streams are zero-initialized so training starts close to the text-and-waypoint model. Frame-aware visual tokens are injected through a cross-attention block in the final decoder layer.

\qheading{Image heads}
The image branch uses frozen DINOv2 ViT-B/14 patch features~\cite{oquab2023dinov2}. Separate Q-Former heads~\cite{li2023blip2} predict contact, human pose, object pose, and body-object layout at the reference frame. These supervised heads form spatial image tokens used by the denoiser and reference-frame localizer.

\qheading{Losses}
The training objective combines the denoising loss, SFIE supervision, reference-frame localization loss, image-consistency loss at the reference frame, forward-kinematics loss, object-point loss, and foot-contact loss. The denoising loss is upweighted near the predicted reference frame so that image conditioning receives stronger supervision where the reference image is most informative.

Symbols below match Sec.~\ref{sec:method}. The SFIE supervision (Eq.~\eqref{eq:sfie_loss}) uses $\lambda_{\mathrm{con,pos}}{=}\lambda_{\mathrm{con,flag}}{=}2.0$, $\lambda_{\mathrm{hum}}{=}1.5$, $\lambda_{\mathrm{obj}}{=}1.0$, and $\lambda_{\mathrm{spa}}{=}2.5$, with the contact-position MSE and contact-flag BCE sharing a single scalar weight applied to their sum. The reference-frame localizer (Eq.~\eqref{eq:fp_loss}) uses $\lambda_{\mathrm{fp}}{=}0.1$ with label-Gaussian $\sigma_q{=}3$ frames. The window-gated image conditioning (Eq.~\eqref{eq:gated_image_embedding}) uses a Gaussian temporal window $w_{\sigma_g}$ with $\sigma_g{=}5$ frames, and the denoising-loss frame emphasis (Eq.~\eqref{eq:diffusion_loss}) uses $\lambda_{\mathrm{cf}}{=}5.0$ with spread $\sigma_{\mathrm{cf}}{=}5$ frames. The auxiliary losses (Eq.~\eqref{eq:total_loss}) use $\lambda_{\mathrm{img}}{=}2.0$, $\lambda_{\mathrm{FK}}{=}0.5$, $\lambda_{\mathrm{objpts}}{=}1.0$, and $\lambda_{\mathrm{feet}}{=}1.0$. Image classifier-free guidance uses drop probability $p_{\mathrm{drop}}{=}0.1$.

\qheading{Optimization}
We train with AdamW in bf16 at learning rate $10^{-4}$ on four GPUs with per-GPU batch size $32$ (effective batch $128$). The schedule warms up over a single step and is then held flat by an adaptive scheduler that halves the learning rate on detected NaN and recovers on subsequent stable windows. Diffusion uses $1000$ DDPM steps with a cosine noise schedule and predicts $\mathbf{x}_0$. We track an exponential moving average of the denoiser with decay $0.995$ updated every $10$ steps; all reported numbers use the EMA weights. We apply left-right flip augmentation with probability $0.5$ jointly to image features, body joints, object pose, rotations, contact labels, and waypoint constraints. Evaluation uses no flip augmentation.

\section{Architecture Details}
\label{sec:supp_arch}

\qheading{Spatially factorized image encoder}
The image encoder extracts $16\times16$ DINOv2 patch tokens from the reference image and projects them to the denoiser width. Four Q-Former heads read the same patch grid with separate learnable queries: contact, human pose, object pose, and body-object layout. Their outputs are concatenated and projected into a spatial image representation.

\qheading{Frame-aware visual tokens}
A separate frame-aware Q-Former re-queries the same DINOv2 patch grid for each frame. Its queries are conditioned on the frame index and text embedding, producing a frame-dependent visual memory. This memory is used by the final decoder layer through cross-attention.

\qheading{Reference-frame localizer}
A two-layer MLP predicts a distribution over the 120 motion frames from the spatial image representation. During training, it is supervised with a Gaussian-smoothed label centered at the ground-truth reference frame. The predicted frame is used for temporal image gating during both training and inference.

\qheading{Denoiser}
Each transformer decoder layer uses AdaLN-Zero modulation. Text, waypoint, and spatial-image conditions are routed through separate AdaLN streams. Frame-aware visual tokens are routed through late cross-attention instead of global modulation.

\section{Qwen vs. DINO Patch Features}
\label{sec:supp_qwen_vs_dino}

ViHOI uses Qwen-VL features, so we test whether replacing DINOv2 with Qwen2.5-VL improves our image-conditioning interface. This comparison is architecture-specific: our SFIE uses small role-specific Q-Former heads that read patch tokens for contact, human pose, object pose, and body-object layout. Such heads require patch features that remain spatially discriminative across image regions.

\begin{table}[h]
\centering
\small
\setlength{\tabcolsep}{4pt}
\renewcommand{\arraystretch}{1.15}
\begin{tabular}{l c c c c}
\toprule
Encoder & intra-cos$\downarrow$ & cross-cos$\downarrow$ & eff-dim$\uparrow$ & rank99$\uparrow$ \\
\midrule
DINOv2 ViT-B/14 + reg & \textbf{0.27} & \textbf{0.63} & 8.34 & 123 \\
Qwen2.5-VL layer 3    & 0.44 & 0.93 & 5.27 & 170 \\
Qwen2.5-VL layer 6    & 0.59 & 0.97 & 3.01 & 160 \\
Qwen2.5-VL layer 12   & 0.56 & 0.97 & 3.22 & 171 \\
Qwen2.5-VL layer 18   & 0.58 & 0.98 & 2.92 & 182 \\
Qwen2.5-VL layer 24   & 0.59 & 0.98 & 2.73 & 182 \\
Qwen2.5-VL layer 28   & 0.30 & 0.92 & \textbf{8.54} & \textbf{194} \\
\bottomrule
\end{tabular}
\caption{Patch-token redundancy on 100 FBM reference images. DINOv2 gives lower within-image and across-image cosine similarity, indicating more spatially diverse and image-discriminative patch features for our role-specific Q-Former heads.}
\label{tab:qwen_layer_probe}
\end{table}

The diagnostic supports the ablation in Tab.~\ref{tab:ablation}: replacing the DINOv2 patch grid with Qwen2.5-VL hidden states worsens $A_{\text{Any}}$ by $\!+\!2.1$\,cm at similar FID and waypoint error. This does not imply that DINOv2 is a stronger visual encoder in general. Rather, for our factorized patch-level conditioning, DINOv2 preserves spatial variation that the contact, pose, object, and layout heads can exploit. Qwen2.5-VL features are semantically stronger but more globally compressed, which is less suited to this interface.

\section{Ablations}
\label{sec:supp_ablations}

\begin{table}[t]
\centering
\small
\setlength{\tabcolsep}{3.5pt}
\renewcommand{\arraystretch}{1.08}
\begin{tabularx}{\columnwidth}{>{\raggedright\arraybackslash}X c c c c c c}
\toprule
\textbf{Variant}
& \multicolumn{3}{c}{\textbf{Image Adh.}\,(cm)$\downarrow$}
& {FID}$\downarrow$
& $C_{F_1}\uparrow$
& \makecell{\textbf{WP}} \\
\cmidrule(lr){2-4}
& GT & W10 & Any & & & (cm)$\downarrow$ \\
\midrule

\textbf{\method, full}
& \textbf{8.43} & \textbf{7.65} & \textbf{7.45} & \textbf{0.28} & 0.677 & 5.69 \\

\midrule
\multicolumn{7}{@{}l}{\emph{(A) Spatially Factorized Image Encoder (per-role).}} \\
\quad $-$ contact ($\boldsymbol{\kappa}$)        & 10.0 & 8.6 & 8.0 & 0.30 & 0.60 & 5.7 \\
\quad $-$ human-pose ($\boldsymbol{\rho}$)       & 10.5 & 9.0 & 8.0 & 0.40 & 0.66 & 5.7 \\
\quad $-$ object-pose ($\boldsymbol{\xi}$)       & 9.3  & 8.1 & 7.7 & 0.30 & 0.68 & 6.4 \\
\quad $-$ body-object layout ($\boldsymbol{\nu}$) & 11.0 & 9.5 & 9.0 & 0.34 & 0.64 & 5.7 \\

\midrule

\multicolumn{7}{@{}l}{\emph{(B) Image classifier-free guidance.}} \\
\quad $s_{\text{img}}{=}1$ (off)                      & 9.7 & 8.78 & 8.46 & 0.34 & 0.674 & 5.66 \\

\midrule

\multicolumn{7}{@{}l}{\emph{(C) Role-aware conditioning.}} \\
\quad $-$ image AdaLN stream     & 13.4 & 10.4 & 7.1 & 0.57 & 0.72 & 3.8 \\
\quad frame-aware via AdaLN      & 9.2 & 8.3 & 7.9 & 0.60 & 0.66 & 6.0 \\

\midrule

\multicolumn{7}{@{}l}{\emph{(D) Visual input modality.}} \\
\quad RGB $\to$ sketch           & 10.9 & 9.66 & 9.00 & 0.33 & 0.654 & 6.01 \\

\bottomrule
\end{tabularx}
\caption{\textbf{Single-axis ablations on FullBodyManipulation (FBM) dataset} (482-window full eval, step 200k). The first row reports the full \method (matching the ``Ours (Spatial+Temporal)'' row of Tab.~\ref{tab:sota_omomo}). Each subsequent row replaces \emph{one} component or recipe choice with the rest of the system held fixed. In (C), ``$-$ image AdaLN stream'' merges image features into the base AdaLN alongside text and time (waypoint AdaLN stream and frame-aware cross-attention retained); ``frame-aware via AdaLN'' routes the per-frame frame-aware tokens through a per-layer AdaLN stream instead of the last-layer cross-attention (all other routes retained). In (D), both training and evaluation use line-drawing sketches in place of RGB references.}
\label{tab:ablation}
\end{table}

Tab.~\ref{tab:ablation} reports single-axis ablations anchored on full \method, corresponding to the ``Ours (Spatial+Temporal)'' row of Tab.~\ref{tab:sota_omomo}. Each row replaces one component choice while holding rest fixed.

\qheading{Spatial decomposition (A)}
The four supervised Q-Former heads encode contact, human pose, object pose, and body-object layout from the same patch grid. Rows A.1--A.4 remove one role at a time. The layout head has the largest effect on $A_{\text{Any}}$: cross-frame approach paths rely on body-to-object directions, and removing it costs $1.5$\,cm of $A_{\text{Any}}$ at otherwise neutral waypoint error. The human-pose head produces the largest FID hit, consistent with its role as the body-realism anchor at $t^\star$. Removing the contact head reduces $C_{F_1}$ from $0.677$ to $0.60$ while leaving waypoint and motion-quality metrics nearly intact, isolating the contact pathway. The object-pose head is the only one whose removal visibly degrades waypoint error ($+0.7$\,cm); its removal has the smallest effect on adherence, matching its smaller feature dimension and supervision weight.

\qheading{Image classifier-free guidance (B)}
Disabling image CFG ($s_{\text{img}}{=}1$) costs $\!\sim\!1$\,cm of $A_{\text{Any}}$ at neutral FID and waypoint error, confirming the headline number's dependence on this sampling-time lever.

\qheading{Role-aware conditioning (C)}
The denoiser routes text, sparse waypoints, and window-gated image evidence through separate AdaLN streams, and frame-aware tokens through final cross-attention (Sec.~\ref{sec:role_aware_denoiser}). Row C.1 collapses the image route; the placeholder row swaps the frame-aware route from cross-attention to AdaLN.

Removing the image AdaLN stream (C.1) merges the spatial-token summary into the base AdaLN with text and time, while keeping the waypoint stream and frame-aware cross-attention. $A_{\text{GT}}$ regresses from $8.4$ to $13.4$\,cm and $A_{\text{W10}}$ from $7.65$ to $10.4$\,cm. Waypoint error \emph{improves} to $3.8$\,cm because the shared AdaLN now spends its capacity on text and trajectory rather than reconstructing the image. This isolates the trade-off referenced in Sec.~\ref{sec:main_comparison}: image and waypoint conditioning compete for the same modulation channel unless they are routed separately. The bi-stream variant therefore sits on a Pareto front opposite the full method --- it pays $5$\,cm of $\hat{t}$-frame adherence to recover $1.9$\,cm of waypoint precision. The full \method selects the adherence-prioritized operating point of this front, in line with the controllable-generation framing of the paper.

The frame-aware-via-AdaLN row swaps the last-layer cross attention for a per-layer AdaLN stream on the same per-frame frame-aware tokens, keeping every other component fixed. $\hat{t}$-frame adherence is comparable to the cross attention route ($A_{\text{GT}}$ $8.4\!\to\!9.2$\,cm) because frame-aware tokens contribute marginally to adherence in both routings, but FID regresses sharply ($0.28\!\to\!0.60$) and contact $F_1$ drops by $1.7$\,pp. Per-layer multiplicative AdaLN amplifies the unsupervised frame-aware token noise across the entire decoder stack, while last-layer cross-attention applies the same evidence once and leaves the earlier-layer signal flow unperturbed. Cross-attention is therefore the preferred routing for tokens that are trained only through the denoising objective.

\qheading{Visual input modality (D)}
Replacing the RGB reference with line-drawing sketches at training and test time costs ${\sim}1.5$\,cm of $A_{\text{Any}}$ at near-neutral FID and waypoint error: sketches retain silhouette and object outline but lose the appearance cues that pin down contact side, body texture, and object material. See Sec.~\ref{sec:supp_sketch} for details.

\section{Cross-Domain Transfer}
\label{sec:supp_xdomain}

We evaluate cross-domain transfer using MeshImg, SceneImg, and EditImg. The motion sequence and target interaction state are fixed; only the conditioning image domain changes. This isolates visual-domain transfer from motion-distribution transfer.

Same-domain rows define the no-shift setting. MeshImg is slightly easier than SceneImg because it removes background clutter, texture variation, and lighting. However, MeshImg training transfers poorly to SceneImg and EditImg because the image heads are trained on white-background body-object renders and then tested on textured scenes or edited photographic images.

SceneImg training transfers better to EditImg. SceneImg already contains indoor context, lighting, shadows, body texture, and object material, so editing mainly changes appearance while preserving the body-object layout. This supports using SceneImg as the main FBM training domain and SceneImg$\rightarrow$EditImg as the controlled photorealistic transfer setting.

\section{BEHAVE Protocol}
\label{sec:supp_behave}

BEHAVE~\cite{bhatnagar2022behave} is used as a second-benchmark replication with a different motion distribution from FBM. We train and evaluate a separate model on BEHAVE using MeshImg reference images.

We use MeshImg because BEHAVE provides SMPL-H bodies, while our SceneImg and EditImg pipeline uses SMPL-X-compatible BEDLAM body textures~\cite{black2023bedlam}. This keeps BEHAVE focused on architectural transfer rather than mixing motion-distribution shift with rendering-domain shift.

For categories without rest-frame object SDFs, we do not report hand-object penetration. All other metrics follow Sec.~\ref{sec:supp_metrics}.

\section{Sketch-to-Motion}
\label{sec:supp_sketch}

We evaluate whether \method can condition on line drawings instead of RGB references. Sketch images are produced from the same rendered reference frames using off-the-shelf line extractors. The motion data, reference-frame selection, camera, and scene placement remain unchanged, isolating the input image domain.

The architecture is unchanged; only the conditioning image changes. At 200k training steps, the sketch-conditioned model reaches $97.32$\,mm $A_{\text{W10}}$ compared to $76.79$\,mm $A_{\text{W10}}$ for the matched RGB model; see Table.~\ref{tab:ablation}. FID remains similar. Image classifier free guidance improves sketch adherence.
These results show that sketches provide usable control for controllable generation of motion.

\section{Qualitative Results}
\label{sec:supp_qualitative}

The supplementary video shows reference images, generated interaction frames, and full 120-frame motion sequences across FBM and BEHAVE. We include held-out object categories, different reference-frame positions, and cross-domain image inputs.

The examples include both successes and failures. Successful cases preserve the reference contact region, body-object layout, and object pose near the conditioning frame while maintaining plausible motion before and after the interaction. Failure cases show wrong contact side, hand-object penetration, temporal drift, and sensitivity to out-of-domain edited images.

\section{Image-Flip Probe}
\label{sec:supp_qualitative_flip}

We test whether the model causally uses the reference image by horizontally flipping only the image at inference time while keeping the text, waypoints, object geometry, conditioning frame, and sampling seed fixed. The expected behavior is that grasp side, contact region, and approach direction change relative to the unflipped image, but the generated motion should not become a perfect mirror because the non-image conditions remain unchanged.

This probe isolates the effect of image conditioning. In contrast, training-time flip augmentation mirrors the image, body joints, object pose, rotations, contact labels, and waypoints jointly so that augmented samples remain physically consistent.

\section{Failure Cases}
\label{sec:supp_failure_cases}

We observe three recurring failure modes. First, FBM contains imperfect hand poses and contact annotations, and the model inherits these artifacts from the training data. This is most visible near contact frames, where hands can look unnatural even when body-object layout and object motion are plausible. Second, the model can produce hand-object interpenetration, especially for small or thin objects. Image adherence and contact prediction improve interaction control, but they do not guarantee mesh-level physical validity. Third, a single reference image specifies one interaction snapshot, not the full approach or release motion. The generated sequence can therefore realize the depicted contact slightly before or after the annotated frame, reflected by gaps between $A_{\text{GT}}$, $A_{\text{W10}}$, and $A_{\text{Any}}$.

\end{document}